\documentclass{acm_proc_article-sp}
\usepackage{tikz}
\usetikzlibrary{calc,positioning,shapes,arrows}
\usepackage{tabularx}
\usepackage{pgfplots}
\usepackage{graphicx}
\usepackage{amsmath}
\usepackage{enumitem}
\usepackage{subcaption}
\usepackage{float}
\usepackage{booktabs}
\usepackage[font=bf]{caption}
\graphicspath{{img/}}
\renewcommand{\footnotesize}{\small}

\begin{document}
\sloppy
\newcolumntype{R}{>{\raggedleft\arraybackslash}X}
\title{A multi-method simulation of a high-frequency bus line using AnyLogic}

\numberofauthors{1} 
%
\author{
%
%
\alignauthor
	Thierry van der Spek\\
	\affaddr{LIACS, Leiden University}\\
	\email{s0922714@umail.leidenuniv.nl}
\alignauthor
	Name Two\\
	\affaddr{LIACS, Leiden University}\\
	\email{sXXXXX@umail.leidenuniv.nl}
}

\permission{This student paper is the result of an academic computer science research project. Permission to make digital or hard copies of all or part of this work for personal or classroom use is granted without fee provided that copies are not made or distributed for profit or commercial advantage and that copies bear this notice on the first page. }

\date{30 July 1999}

\maketitle
\begin{abstract}
\label{sec:abstract}	
In this work a mixed agent-based and discrete event simulation model is developed for a high frequency bus route in the Netherlands.  
With this model, different passenger growth scenarios can be easily evaluated. This simulation model helps policy makers to predict changes that have to be made to bus routes and planned travel times before problems occur. 
The model is validated using several performance indicators, showing that under some model assumptions, it can realistically simulate real-life situations. The simulation's workings are illustrated by two use cases.
\end{abstract}

\section{Introduction}
\label{sec:introduction}
In all metropolitan areas, people rely on buses as a means of transportation. Commuters increasingly use public transport to get to or go from their work or study location in these areas. This puts a high strain on bus companies to provide a reliable and frequent  service, especially during rush hours. The procession of buses is easily disrupted by different factors such as congestion, traffic lights, open bridges and passengers boarding or alighting. This causes delays, which again may cause bus \textit{bunching}. Bus bunching, the uneven spacing of buses over a route, causes uneven loading of passengers and therefore inefficient usage of material. Furthermore, passenger satisfaction may be lower due to full buses. 

In the \textit{Randstad} megalopolis in central-western Netherlands, several high-frequency bus lines are created to interconnect cities. Such a high-frequency line provides a quick connection between places, but inherently is challenged by these disruptive influences. This work aims to create insight into these factors and their effect on travel time on one of these high-frequency bus lines between the cities of Leiden and Zoetermeer, exploited by the Arriva bus company.

We show that a combination of \textit{agent-based} and \textit{discrete-event} models using the AnyLogic simulation tool\footnote{Anylogic~Multimethod~Simulation, http://anylogic.com/, Visited: Feb 17, 2017} are a suitable approach to visualize this bus line and create a better comprehension of frequently occurring problems. One of the external factors, the number of passengers, is modelled and can be configured to analyse several passenger growth scenarios. The effect of these scenarios is illustrated by several key performance indicators (KPIs), determined by Arriva.

This paper is structured as follows: In Section~\ref{sec:domain}, common domain terminology is introduced to gain a better understanding of the problems occurring. In Section~\ref{sec:problemdefinition}, the problems occurring are described by formulating our KPIs. In Section~\ref{sec:relatedwork}, we look into related work in this domain, and also at other agent-based and discrete-event simulation models. Section~\ref{sec:datasets} discusses the data that is considered and the data sets that are used. Also, the preprocessing of data that is needed to generate our model is discussed. Thereafter, in Section~\ref{sec:approach}, the modelling choices that are made are discussed. In Section~\ref{sec:experiments}, the discussed model is validated using the defined KPIs and two use cases of passenger growth are discussed. Finally, in Section~\ref{sec:conclusions}, we conclude this work and possible future perspectives are briefly discussed.

\section{Background Information}
\label{sec:domain}
In this section, we go over some basic definitions that are needed to gain a better understanding of the problem definition and the proposed model. Furthermore, common characteristics for service quality and reliability that the proposed model provides insight into are defined.

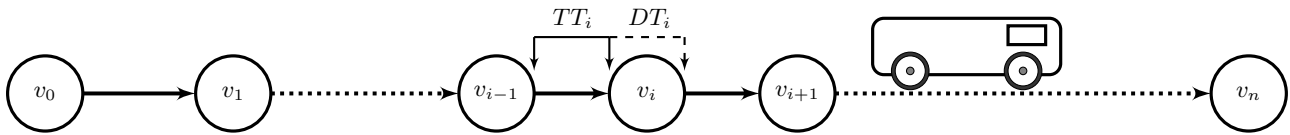
\begin{figure*}[ht!]
	\centering
	\begin{tikzpicture}
	
	\node[draw,circle,minimum size=1cm,inner sep=0pt,very thick] at (0,2.5) {$v_0$};
	\draw[-latex', line width=1.5] (0.5,2.5) -- (2,2.5);
	
	\node[draw,circle,minimum size=1cm,inner sep=0pt,very thick] at (2.5,2.5) {$v_1$};
	\draw[-latex', dotted, line width=1.5] (3,2.5) -- (5.5,2.5);
	
	\node[draw,circle,minimum size=1cm,inner sep=0pt,very thick] at (6,2.5) {$v_{i-1}$};
	\draw[-latex', line width=1.5] (6.5,2.5) -- (7.5,2.5);
	
	\node[draw,circle,minimum size=1cm,inner sep=0pt,very thick] at (8,2.5) {$v_i$};
	\draw[-latex', line width=1.5] (8.5,2.5) -- (9.5,2.5);
	
	\node[draw,circle,minimum size=1cm,inner sep=0pt,very thick] at (10,2.5) {$v_{i+1}$};
	\draw[-latex', dotted, line width=1.5] (10.5,2.5) -- (15.5,2.5);
	
	\node[draw,circle,minimum size=1cm,inner sep=0pt,very thick] at (16,2.5) {$v_n$};
	
	\filldraw[fill=white,rounded corners=1.1ex, draw=black,very thick] (11,2.75) rectangle (13.5,3.5);

	\filldraw[fill=white,draw=black,very thick] (12.8,3.15) rectangle (13.3,3.4);
	
	\draw[draw=black,fill=black!80] (11.5,2.8) circle (.25);
	\draw[draw=black,fill=white!50] (11.5,2.8) circle (.2);   
	\draw[draw=black,fill=black!50] (11.5,2.8) circle (.05);  
	
	\draw[draw=black,fill=black!80] (13,2.8) circle (.25);
	\draw[draw=black,fill=white!50] (13,2.8) circle (.2); 
	\draw[draw=black,fill=black!50] (13,2.8) circle (.05);
	
	\draw[-, thick] (6.5,3.25) -- (7.5,3.25);
	\draw[-, thick, dashed] (7.5,3.25) -- (8.5,3.25);
	\draw[-latex', thick, dashed] (8.5,3.25) -- (8.5,2.8);
	\draw[-latex', thick] (6.5,3.25) -- (6.5,2.8);	
	\draw[-latex', thick] (7.5,3.25) -- (7.5,2.8);
	\node[text width=1cm] at (7.25,3.5) {$TT_i$};
	\node[text width=1cm] at (8.25,3.5) {$DT_i$};	
	\end{tikzpicture}
	\caption{Schematic overview of bus stops along one direction of a bus line with $n+1$ stops. The arrow indicates the direction in which the bus is travelling. Dots indicate possible other stops on the route. Stops $0$ and $n$ are terminal stops.}
	\label{fig:tripoverview} 
\end{figure*}

\subsection{Preliminaries}
In this section, notation needed to define the bus route and terminology used in the field of public transport are given. Some definitions might be common knowledge, but others are less known outside this discipline. 

A group of bus lines can be seen as a graph $G = (V,E)$, which is a combination of a set of \textit{stops} $V = \{v_0,v_1,v_2,\dots,v_n\}$ and a set of route \textit{segments} $E \subseteq V \times V$. A stop $v_i$ is a designated place at which passengers can alight or board the bus. 

We define $P$ to be the set of all possible bus trips: 
\begin{equation}
\begin{split}
P = \{v_0,\dots,v_{n} \mid \forall 0 \leq i < n : 
(v_{i},v_{i+1}) \in E,\\ 
\forall 0 \leq h < i \leq n : v_h = v_i \implies (h,i) = (0,n) \}
\end{split}
\end{equation}

A bus trip $p \in P$ of length $n$ can then be seen as an ordered sequence of $n+1$ stops, where there exists a route segment $e_i \in E$ from every stop $v_{i}$ to its subsequent stop $v_{i+1}$.

We call stops $v_0$ and $v_n$ \textit{terminals}. At these stops, the bus changes trips.
Figure~\ref{fig:tripoverview} shows a schematic overview of a trip.

Between stops, the bus drives these specific route segments. The time that each segment $e_i$ takes is called \textit{travel time} ($TT_i$). Travel time can be measured between stop departures, between departure at stop $v_i$ and arrival at stop $v_{i+1}$ or vice versa. We define travel time ($TT_i$) between stop $v_{i-1}$ and $v_i$ as the time between departure at stop $v_{i-1}$ and arrival at stop $v_i$, which is the time it takes to travel segment $e_i$. We note that when we talk about travel time, we talk about the time that it takes to travel a route segment towards a stop from its preceding stop, therefore $e_0$ is not defined.

Bus lines can either consist of a single trip in which the start terminal is the same as the end terminal $v_0 = v_n$, or two trips in roughly opposite directions $v_0 \neq v_n$. For the first case we have a circular bus line, which is more common in urban areas. In the latter case most bus stops are usually in opposite sides of the road to allow passengers to travel in both directions of the route. 

Furthermore, stops are defined by a GPS radius. This GPS radius is roughly 35 metres. The time the bus spends within the radius of stop $v_i$ is known as \textit{dwell time} of stop $v_i$ ($DT_i$). Within this GPS radius the bus stops to let passengers board or alight the bus. Doors open within a stop's GPS radius. The time that the bus' doors are open within the radius of stop $v_i$ is defined as \textit{door open time} ($DOT_i$).
 
Bus operators have a specific itinerary of trips that they drive during the day, this is called the operators' \textit{run} $R \subset P$. A run implies that operators possibly have to switch buses at terminals. Buses also have a specific itinerary of trips $C \subset P$ which called the bus' \textit{circulation}. A bus circulation could be on the same line, but a bus may also switch lines, thus changing routes.
 
Buses can also enter or leave the system at terminals. Some parts of the day (off-peak hours) may have longer \textit{headways}, which is the time between one bus and the next bus on the same trip. This means total bus capacity is reduced and some buses return to the garage to be stored. The drive from and to the garage is not indicated as a trip, but is called \textit{deadhead}. This happens at the start and end of a bus' circulation. A bus can have multiple circulations during the day thus also having multiple deadheads.

\section{Problem definition}
\label{sec:problemdefinition}
The simulated R-net 400 line has 12 buses that each are driven along 2 specific trips. These trips are south-bound (A-direction) from the Leiden terminal to Zoetermeer and north-bound (B-direction) in the opposite direction. Each trip consists of 11 stops, including terminals. Deadheads are only made from and to the Leiden terminal. The buses' circulations are closed for this bus line, meaning that only these two trips are made by these 12 buses. This is because they are part of the R-net concept\footnote{R-net~concept, http://www.rnet.nl/corporate/, Visited: Feb 17, 2017}, which only allows these specific buses to drive designated routes.

To create a reliable bus line there are some important factors that have to be optimized. In this work, the R-net 400 bus line is simulated, whilst giving insight into some of these factors. Two main components of customer satisfaction are waiting time and seat availability~\cite{vanoort2011service}. We discuss both of them below. 

\paragraph{Waiting time}
Waiting time is mainly determined by bus \textit{punctuality} and \textit{regularity}.

\begin{itemize}
	\item \textbf{Punctuality}: punctuality is usually seen as being on time. Punctuality shifts are caused by any deviation from the scheduled arrival time at a stop. In this paper, the term punctuality is used as the inverse, so that an increase in punctuality defines a larger deviation from the schedule. This can be calculated for a single stop, or all stops of a line. For a stop $v_i$, we can for example calculate the average (departure) punctuality $\bar{Q_i}$ on a trip $p$ as follows:
	\vspace{0.25cm}\\
	\begin{equation}
	\bar{Q_i} = \frac{
	\sum\limits_{j=1}^{k_i} 
	D_{i,j}^{\textsc{act}} - D_{i,j}^{\textsc{sched}}
}{k_i}
	\label{eq:punctualityavg}
	\end{equation}
	where:\vspace{0.1cm}\\
	$k_i$: number of buses departing from stop $i$\vspace{0.1cm}\\
	$D_{i,j}^{\textsc{sched}}$: scheduled departure time at stop $i$ of vehicle $j$\vspace{0.1cm}\vspace{0.1cm}\\
	$D_{i,j}^{\textsc{act}}$: actual departure time at stop $i$ of vehicle $j$\vspace{0.1cm}\vspace{0.1cm}\\
	
	The formulation in Equation~\ref{eq:punctualityavg} has the downside that it does not indicate if vehicles depart too early or too late, but compensates departures that are too early with departures that are too late. One solution is to take the absolute value of the difference, but then trips being too early would count as being too late. 
	
	\item \textbf{Regularity}: irregularity is determined by any deviation from the scheduled headway. Since measurements are taken at bus stops, regularity can be calculated between bus stop departures. For a specific bus stop $v_i$ the average deviation from the scheduled headway $\bar{R_i}$ can be calculated as follows:\vspace{0.25cm}\\
	\begin{equation}
	\bar{R_i} = \frac{
		\sum\limits_{j=1}^{k_i} 
		\cfrac{
			H_{i,j}^{\textsc{act}} - H_{i,j}^{\textsc{sched}}
		}{H_{i,j}^{\textsc{sched}}}
	}{k_i}
	\label{eq:regularityavg}
	\end{equation}

	where:\vspace{0.1cm}\\
	$k_i$: number of buses departing from stop $i$\vspace{0.1cm}\\
	$H_{i,j}^{\textsc{sched}}$: scheduled headway time at stop $i$ of vehicle $j$\vspace{0.1cm}\vspace{0.1cm}\\
	$D_{i,j}^{\textsc{act}}$: actual headway at stop $i$ of vehicle $j$\vspace{0.1cm}\vspace{0.1cm}\\

\end{itemize}
These measures focus on the supply side, with the assumption that there is a uniform arrival pattern of travellers. We return to this assumption later when we discuss our model assumptions.

\paragraph{Seat availability} 
Seat availability is determined by the net result of people boarding and alighting the bus. We can calculate the average occupancy of the bus $\bar{O_i}$ after the bus left a stop $i$ as follows:\vspace{0.25cm}\\

\begin{equation}
\bar{O_i} = \cfrac{\sum\limits_{j=1}^{k_i}\sum\limits_{x=0}^{i}P_{x,j}^{\textsc{OUT}} - P_{x,j}^{\textsc{IN}}}{k_i}
\label{eq:seatavailability}
\end{equation}
where:\vspace{0.1cm}\\
$k_i$: number of buses departing from stop $i$\vspace{0.1cm}\\
$P_{x,j}^{\textsc{IN}}$: number of passengers boarding at stop $x$ into bus $j$\vspace{0.1cm}\\
$P_{x,j}^{\textsc{OUT}}$: number of passengers alighting at stop $x$ from bus $j$
\vspace{0.1cm}\vspace{0.1cm}\\

All measures are determined by travel time, dwell time and the number and type of buses. We return to these measures later, when formulating our simulation model.

The issue of optimizing customer satisfaction is a manifold problem involving the optimization of waiting time, and seat availability. This means that punctuality and regularity need to be optimized under the constraints of seat availability, the number of buses, travel time and dwell time. These constraints are correlated and again dependent on external factors such as number of passengers and traffic conditions. By giving insight into the defined KPIs we show that we create an understanding of this complex process.

\section{Related and previous work}
\label{sec:relatedwork}
In this section, previous work that has been done on discrete-event simulation and agent-based modelling in public transport is discussed. Also, the combination of both modelling approaches in related domains is given.

\subsection{Discrete-event models}
Discrete event simulation models the operations of a system as a sequence of discrete events at which the state of the system changes~\cite{robinson2014simulation}. Each event occurs at a specific instant in time and marks a change in the state of the system. Between events, no change in the state of the system is assumed, so the simulation clock jumps to the time of the next event. 

A discrete-event simulation for dynamic transit operations has been introduced in~\cite{toledo2010meso}. This model is built on top of a mesoscopic traffic simulation, which allows modelling of operation dynamics. It is used to design control strategies for high-frequency bus lines such as in our work. One downside of their approach is the lack of a definition of microscopic behaviour and a centralized approach. Global behaviour is defined by flowcharts, typically with stochastic elements~\cite{borshchev2004system}.

The difference with agent-based simulation is that micro-level entities have no `intelligence' so their aggregate behaviour is pre-defined. Our model incorporates this global specified behaviour of the system, but also combines this with decentralised specified behaviour of individual actors within the system.

\subsection{Agent-based models}
Agent-based computing is a computational model, simulating the behaviour of autonomous decision making agents~\cite{bonabeau2002agent}. The usage of agent-based models requires a decentralised nature of the problem, where each agent behaves according to their objectives within their particular environment, based on input from that environment~\cite{jennings1999agent}. These behaviours usually follow simple rules, but behaviour can be specified to be more complex as well. This decentralised approach differs from the centralized approach of discrete-event simulation. There is no definition of global system behaviour, but it emerges from the actions and interactions of agents. 

Agent-based modelling is used previously to model passengers' behaviour, when choosing for public transport and the effect of infrastructure investments and policy changes~\cite{hajinasab2016agenttransport}. Here agents base their individual decisions on cost, time, convenience and social norm. The simulation investigates passenger departure time and choice of transport. This model, however, is mostly passenger based and investigates their choice for public transport, where our method looks into the effect on one specific bus line.

Furthermore, agent-based models are used to examine the effect of bus rapid transit measures such as bus lanes and bus priority systems~\cite{mcdonnell2011exploring}. A comparison of design measurements is made under static passenger amounts. It is shown that several design measurements decrease passenger waiting time and travel time. Our work, however, focusses more on the dynamic passenger numbers in a static environment.

\subsection{Multi-model approaches in other domains}
The combination of discrete-event simulation and agent-based modelling is created in our work using the Anylogic tool. With this tool, several models have been created in different fields, for example for the analysis of power and performance of data centers~\cite{postema2015anylogic}, the outbreak response of immunization~\cite{doroshenko2016evaluation} and airport checkpoint pedestrian flow~\cite{jung2016creating}.

In these systems, using discrete-event simulation, the evolution of the changing system over time is captured. At the points in time where events occur, behaviour is defined by individual agent behaviour. This creates a hybrid model. These models show that capturing the complexity of a real-life situation in a model, whilst keeping model simplicity, sometimes requires a combination of different paradigms. Therefore this multi-model approach is also used in our work.\\

\section{Data sets and pre-processing}
\label{sec:datasets}
Before going into detail about the model, it is useful to first describe the data that is used to create the model. This will give some insight into some of the model restrictions and assumptions that are made. Additionally, we describe the steps that are made to preprocess the data sources used. 

\subsection{Data sets}
We look into several internal and external data sources.  Most data sources influence travel time or dwell time directly or indirectly. Not all data sources are used in the model as some are not suitable or incomplete. Each of the considered data sources is discussed and stated whether it is used.

\subsubsection{Internal data sources}
For the simulation, three weeks of data from the R-net 400 bus line is used. According to domain experts, these weeks are identified as standard weeks, so these weeks have normal passenger behaviour, no detours or other rare occurrences on this bus line. As mentioned, the bus line is a closed system of 12 buses with their own circulations. 
\begin{description}[style=nextline,leftmargin=1em]
\item[Bus log]
Initial bus logging data is provided as Excel data files, which contain historic events logged by the buses' information systems.
Several events are logged by these bus logging systems of which the most important ones are listed in Table~\ref{tab:datasetoverview} with their associated data fields.

\begin{table}[h!]
	\small
	\centering
	\caption{Description of the main events in the bus log data set. Each bus logs several events each with several data fields. The most important are listed here.\vspace{0.5em}}
	\label{tab:datasetoverview}
	\begin{tabularx}{\linewidth}{lll}
		\hline
		
		\textbf{event} & \textbf{data fields} & \textbf{description} \\\hline
		\begin{tabular}[t]{@{}l@{}}stop arrival/\\ departure\end{tabular} & \begin{tabular}[t]{@{}l@{}}datetime\\ coordinates \\ bus id \\trip id \\ driver id \\ stop id\end{tabular} & \begin{tabular}[t]{@{}l@{}}the arrival or departure\\ at a stop identified by \\ the bus' GPS system.\end{tabular} \vspace{1em}\\
		\begin{tabular}[t]{@{}l@{}}door open/\\ closed\end{tabular}       & \begin{tabular}[t]{@{}l@{}}datetime\\ coordinates \\ bus id \\ trip id \\ driver id \\ stop id\end{tabular} & the door opened or closed                                        \vspace{1em}\\
		KAR in/out                                                        & \begin{tabular}[t]{@{}l@{}}datetime\\ coordinates \\ bus id\end{tabular}                                                                          & \begin{tabular}[t]{@{}l@{}}traffic light communication\\  in and out signal \\ (to and from the traffic light)\end{tabular}
	\end{tabularx}
\end{table}

Each of the events are described in more detail:

\begin{itemize}
\item A stop arrival and departure is logged at the point the bus enters and leaves the stop's GPS radius. A trip id combined with the date part of the datetime is a unique identifier for that trip. For the A-direction, trip id's start at 1001 and each subsequent trip in the A-direction for this route gets a trip id, which is the next odd number. So the second trip in that direction of the day will get the number 1003. B-direction trips start with 1002 and each subsequent trip is the next even number. 

\item The door open and door close events are within the time that a bus is at the GPS radius of a stop. These events are also linked to the trip by the trip id and date.

\item The KAR event requires somewhat more explanation. All traffic lights on the route this bus travels are equipped with a short distance radio (Korte Afstands Radio (KAR) in Dutch). As a bus approaches a traffic light, a radio signal is send from the bus to the traffic light and a ``message received" acknowledgement is returned from the traffic light. At that point the traffic light system tries to give priority to the bus, based on some rules set by traffic light policy makers. After the bus passed the traffic light, another message is send from the bus and returned from the traffic light to indicate that the bus passed the traffic light.

\end{itemize}
Due to a timely data export function, we are restricted to a rather small data set of 15 working days. One of the assumptions, also based on expert knowledge, is that this data represents the situation in other weeks as well. To check if there are differences between the days in the data set we used ANOVA~\cite{ostle1975statistics} on dwell time and travel time to see if there are significantly different days. We did not find any significant differences on these two measures. We do note that these data sets only contain week days as weekend days have different schedules/circulations. Week days most likely do differ significantly from weekend days. Weekend days are however less interesting for our model as these days are most likely not when most problems will occur.

We use most of the bus log data source. However, we do not consider KAR events. After plotting the average KAR time for each of the traffic lights on the route and showing them to a domain expert, the conclusion is that there is not much to optimize on this route.

	\item[Passenger data] 
	Passenger data is obtained from the automatic passenger check-in and check-out systems that are in every bus. When entering the bus, passengers check-in and when alighting, they check-out. This data is separately stored, but is linked to a trip id. This data source is used in our simulation and combined with the other data sources.
	\item[Bus schedules]
	Another internal data source that is used are the bus schedules. Bus schedules are a table of trip ids and stop ids. These tables indicate at which stop and time a bus should be.
	\item[Bus circulations]
	Furthermore, a list of bus circulations is used. These circulations are lists of trip ids for a specific bus id, indicating which trips a bus should drive during the day.
\end{description}

\subsubsection{External data sources}
There are many external factors that determine travel time in buses. Factors like congestion, road works, open bridges and even weather all affect travel time. Some of these factors might even be highly correlated, like road works and congestion. 

\begin{description}[style=nextline,leftmargin=1em]
	\item[Road works] 
	When road works occur, the bus company is informed and buses are possibly redirected. Stops may be skipped or the whole trip might change. This is a very complex situation and not happening in our data set and therefore we choose to omit this factor. 
	
	\item[Congestion]
	Actual congestion information is available through the National Database of Road Traffic\footnote{National Databank Wegverkeergegevens, http://www.ndw.nu, Visited: Feb 21, 2017} (NDRT), but unfortunately this is mainly focussed on highways and most of our bus' route is not covered. There is some historical congestion information based on cellphone data at Google, but unfortunately this data is not available for third parties. This is why this data source also is omitted.
	
	\item[Open bridges]
	On the studied route there is a bridge. Open bridges can be of significant influence on travel time. In the NDRT there is also information available about open bridges. The bridge on our route is known to frequently open during rush hour, however, unfortunately the bridge detection system was not properly working in recent months and data is not available. This data source is therefore also omitted.
\end{description}

\noindent By looking at available data sources, we can conclude that we do not have enough determining sources to predict travel time, but we can look at dwell time prediction as we assume this is mainly caused by passengers boarding and alighting the bus at stops. Therefore, in our simulation, we used travel time as a given factor and sample it from historical data. How travel time is obtained is described next.

\subsection{Pre-processing}
\begin{figure}[h!]
	\centering
	\begin{tikzpicture}
	\tikzset{
		mynode/.style={rectangle,rounded corners,draw=black, very thick, inner sep=0.5em, minimum height=2em,minimum width=7em, text centered},
		database/.style={
			cylinder,
			cylinder uses custom fill,
			shape border rotate=90,
			aspect=0.25,
			minimum height=5em,
			draw,
			very thick
		}
	}  
	\node[draw,mynode] at (0,0) {circulations};
	\node[draw,mynode] at (0,1) {bus log};
	\node[draw,mynode] at (0,2) {schedule};
	\node[draw,mynode] at (0,3) {passenger log};
	\node[draw,mynode] at (3.5,1.5) {processing};
	\node[draw,database] at (6,1.5) {total log};
	
	\draw[-latex', thick] (3.5em,0em) -- (7.3em,3.8em);
	\draw[-latex', thick] (3.5em,3em) -- (7.2em,4.3em);
	\draw[-latex', thick] (3.5em,6em) -- (7.2em,4.8em);
	\draw[-latex', thick] (3.5em,9.5em) -- (7.2em,5.3em);
	
	\draw[-latex', thick] (14.25em,4.75em) -- (16.5em,4.75em);
	\end{tikzpicture}
	\caption{Passenger check-in/out data is combined with the bus schedule, the bus logging system and the bus circulations. An overview per trip of events is created. Punctuality, door open time, dwell time, travel time are calculated in this processing step. The combined data is stored by trip in a database called the total log.}
	\label{fig:processingsteps} 
\end{figure}
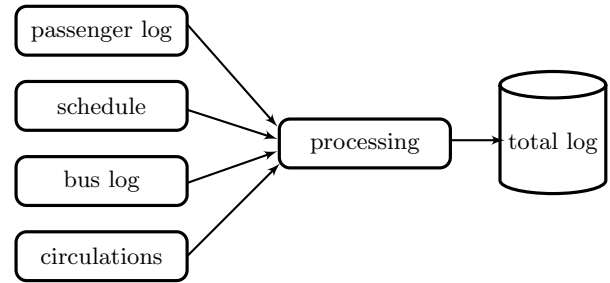

The used data sources are each parsed in their own way and combined to a total log containing all information about each trip. The Excel files containing the passenger check-ins and check-outs are parsed to extract counts for each of the stops. This is combined with information from the bus log for when the bus is at a stop. 
From the bus log, door open time per stop, dwell time per stop and travel time per stop are calculated. Door time is easily calculated as the difference between the door open and close events. Dwell time and travel time can be calculated from arrival and departure times at stops.
Coordinates of the bus log are converted from the used Rijksdriehoek coordinate system (AME-7) to our simulation coordinate system World Geodetic System 1984 (WGS-84) so we can plot them later in our simulation. 
Additionally, the data is combined with the bus schedule to calculate punctuality, and circulation id's are added.
The global overview of the processed data sources is given in Figure~\ref{fig:processingsteps}.

The total log database contains trip objects, each consisting of the following data fields:
\begin{itemize}
	\item datetime
	\item trip id
	\item bus id
	\item circulation id
	\item driver id
	\item per stop:
	\begin{itemize}
		\item coordinates of the stop
		\item travel time to stop (departure previous stop - arrival current stop)
		\item dwell time (arrival current stop - departure current stop)
		\item door open time (door closed time - door open time)
		\item punctuality (departure current stop - scheduled departure)
		\item check-ins and check-outs
	\end{itemize}
\end{itemize}
\vspace{1em}

\noindent As shown in Section~\ref{sec:approach} that this data is sufficient to start creating our model. Unfortunately we have to deal with some missing data as well. After inspecting the data, around 11\% of all trips had one of the following errors: missing stop arrival or departure data point, missing door open or door closed event, whilst check-ins or check-outs occur, or incorrect timestamps. In some cases, departure happened before arrival. These incorrect data points are omitted and not used for creating the model.

\section{Approach}
\label{sec:approach}
In this section, the modelling choices made and how they are implemented are described. A global overview of the model is given first. Secondly, the modelling of passenger growth is discussed. Thereafter, the visualization of the model is explained and some of the model assumptions are discussed.

\subsection{Model overview}
We start out by creating a model for simulating a regular working day. Bus routes are drawn on a map using the stop coordinates of the specific trips and defined segments in between them. This simulation is set to run a working day, starting at 5:00 AM, ending at 1:00 AM the next day. Within this time frame, all daily trips on this line take place.

Buses are generated with discrete-event simulation. Generation time is based on the bus' circulation. Buses are placed at the coordinates of the first stop, so the deadhead is not shown. 

Passengers are then generated by the discrete-event simulation. We know the distribution of passengers that have travelled with specific trips from our data. On high-frequency bus lines such as ours, passengers arrival rates usually follow the Poisson distribution~\cite{fu2002design,dessouky2003real}. We have therefore modelled the distribution of passenger arrival rates as in Equation~\ref{eq:poissonarrival}.
\vspace{0.25cm}\\
\begin{equation}
B_{ij} \sim \cfrac{\lambda_{ij}^k e^{-\lambda_{ij}}}{k!}
\label{eq:poissonarrival}
\end{equation}
where:\vspace{0.1cm}\\
$B_{ij}$: the number of passengers boarding at stop $i$ for trip $j$\vspace{0.1cm}\\
$\lambda_{ij}$: the arrival rate at stop $i$ within the scheduled headway of trip $j$\vspace{0.1cm}\\
$k$: the number of arrivals, $k\in\mathbb{N}$\vspace{0.1cm}\\

Buses and passengers are generated as agents and act according their own set of rules. These rules are specified next.

\subsubsection{Agents}
The two main type of agents that are defined are buses and passengers. However, for convenience, a bus stop agent type is also created. As stated in Section~\ref{sec:relatedwork}, an agent's behaviour can be as simple or complex as we define it to be. This allows to create helper agents, such as bus stops, to interact with other agents. 
Agent behaviour is specified using state chart diagrams, resembling the real life behaviour of the objects represented. Each state and transition in these diagrams is additionally programmed to follow certain behavioural rules. These rules are described in more detail in the following.

\begin{figure}[h!]
	\centering
	\begin{tikzpicture}[auto]
	\tikzset{
		state/.style={rectangle,rounded corners,draw=black, very thick, inner sep=0.5em, minimum height=2em,minimum width=7em, text centered},
		choice/.style={rectangle,draw=black, very thick, inner sep=0.5em, minimum size=1em, text centered,rotate=45},
		line/.style ={draw, thick, -latex', shorten >=0pt},
		enter/.style={draw, thick, -latex',  shorten >=0pt}
	}  	

	\node[draw,state] (unloading) at (1.75,6)  {unloading};
	\node[draw,state] (holding) at (1.75,3)  {holding};
	\node[draw,state] (loading) at (1.75,1.5)  {loading};
	\node[draw,state] (moving) at (1.75,0)  {moving};
	\node[choice] (choicepoint) at (1.75,4.5) {};

	\draw[black,fill=black] (0.1,3) circle (.1);
	\draw[-latex', very thick] (0,3) -- (0.6,3);
	\draw[-, very thick] (0,2.8) -- (0,3.2);
	
	\draw[->, very thick, dashed]  (1.55,4.5) -- (0.5,4.5);
	\draw[black,fill=white] (0.3,4.5) circle (.2);
	\draw[black,fill=black] (0.3,4.5) circle (.1);
		
	\path[line] (unloading)   -- (choicepoint);
	\path[line] (choicepoint) -- (holding);
	\path[line] (holding)     -- (loading);
	\path[line] (loading)     -- (moving);
	\draw[-, thick] (2.9,0) -| (3.5,6);
	\draw[-latex', thick] (3.5,6) -- (2.9,6);
	
	\end{tikzpicture}
	
	\caption{Bus agent state chart diagram. Holding indicates the bus is waiting without passengers until it is scheduled to departure. (Un)loading is the process of passengers alighting or boarding the bus. The moving state means the bus is moving from one stop to the other.}
	\label{fig:busagentstatechart}
\end{figure}
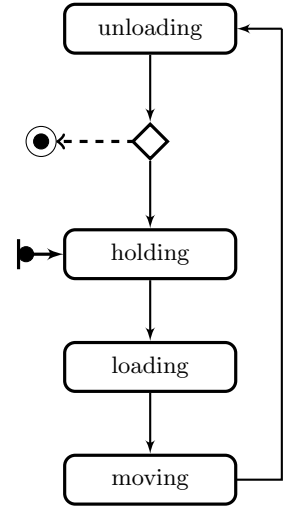

\paragraph{Bus agent}
After bus agents are generated, bus agent's behaviour is specified in the form of a state chart diagram in Figure~\ref{fig:busagentstatechart}.

The blocks are programmed to the following rules:
\begin{itemize}
	\item \textbf{Holding}: if the bus is at the initial stop of the trip, the bus is held until the scheduled departure time of that trip. After that, from the historical departure distribution for that trip, a value is drawn. The bus will then depart, according to that drawn departure time, from the first stop. If the bus is at any other stop, the holding state is not used.
	
	\item \textbf{(Un)loading}: if not at the initial stop, the bus, depending on passengers, will simultaneously load and unload them. At the first stop, the bus will only load passengers and at the last stop, the bus will only unload. 
	In these steps, the bus agent interacts with passenger agents. A message is first send to passengers currently travelling on the bus. If the stop is at any of the passengers' destination, the bus will unload them.
	
	A message is also send to the passenger agents waiting at the specific stop. This is not done directly, but through the mentioned helper agent, the stop agent. Based on the number of passengers waiting at the stop and the capacity left after unloading, the bus will send a message to the stop agent, containing the number of passengers that can be loaded. Then, the stop agent sends a message to the passenger agents at the stop, telling them to board.
	
	The time it takes to load and unload passengers is determined by a linear model. The determination of the coefficients of this model is discussed later in this section.
	
	\item \textbf{Moving}: after loading and unloading, the bus is moving from one stop to the next. The time this takes is drawn from the historical distribution of travel times for that specific trip. Just before departing, the bus will send a message to all passengers that are travelling on the bus, so that passengers that boarded can change states.
	
\end{itemize}

\begin{figure}[h!]
	\centering
	\begin{tikzpicture}[auto]
	\tikzset{
		state/.style={rectangle,rounded corners,draw=black, very thick, inner sep=0.5em, minimum height=2em,minimum width=7em, text centered},
		choice/.style={rectangle,draw=black, very thick, inner sep=0.5em, minimum size=1em, text centered,rotate=45},
		line/.style ={draw, thick, -latex', shorten >=0pt},
		enter/.style={draw, thick, -latex',  shorten >=0pt}
	}  	
	
	\node[draw,state] (waiting) at (0.2,7.5)  {waiting};
	\node[draw,state] (boarding) at (0.2,6)  {boarding};
	\node[draw,state] (travelling) at (0.2,4.5)  {travelling};
	\node[draw,state] (alighting) at (0.2,3)  {alighting};
	\node[draw,state] (arrived) at (0.2,1.5)  {arrived};
	
	\draw[black,fill=black] (0.2,8.5) circle (.1);
	\draw[-latex', very thick] (.2,8.6) -- (0.2,7.8);
	\draw[-, very thick] (0,8.6) -- (0.4,8.6);
	
	\draw[->, very thick, dashed]  (0.2,1.15) -- (0.2,0.2);
	\draw[black,fill=white] (0.2,0) circle (.2);
	\draw[black,fill=black] (0.2,0) circle (.1);

	\path[line] (waiting)    -- (boarding);
	\path[line] (boarding)   -- (travelling);
	\path[line] (travelling) -- (alighting);
	\path[line] (alighting)  -- (arrived);	
	\end{tikzpicture}
	
	\caption{Passenger agent state chart diagram. }
	\label{fig:passengeragentstatechart}
\end{figure}
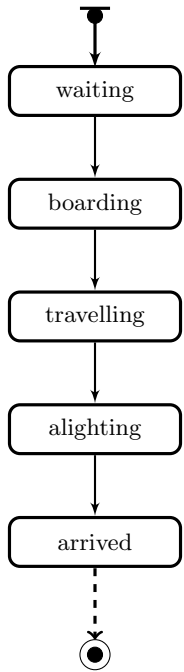

\paragraph{Passenger agent} 
Passenger agents are the other main agent type in the model. Passengers are generated and have their origin and destination stored. The point at which they appear on the GIS map is at their stop of origin. Passenger behaviour is described in Figure~\ref{fig:passengeragentstatechart}. 

\vspace{1em}
The following states are identified:
\begin{itemize}
	\item \textbf{Waiting}: The passenger is waiting at a stop. At this point the passenger is waiting until the next bus arrives at that stop. Since multiple passengers may wait for the bus, a FIFO queue of passengers is held. This is actually done by a helper agent, called the stop agent. Its main action is to queue passengers and to release a number of passengers from the queue. This is done after a message is received from a bus agent, that it arrived at that stop.
	\item \textbf{Boarding}: When the bus arrives, the queue is (partly) emptied by the stop agent and the passenger is boarding the bus. 
	\item \textbf{Travelling}: When the last passenger at the stop is loaded or the bus is full, all passengers on the bus get a message from the bus that it enters the moving state. This is a signal for the passenger to move to the travelling state.
	\item \textbf{Alighting}: Once the bus has arrived at a stop, all passengers on it receive a message from the bus that it has arrived and if this is a passenger's destination stop, the passenger moves to the alighting state.
	\item \textbf{Arrived}: After the passenger has alighted, it is at its destination and can leave the system.
\end{itemize}
\vspace{1em}

\subsection{Passenger growth}
To simulate passenger growth, the effect of passenger growth on total time travelled has te be determined. Total travel time between terminals is the sum of travel times and dwell times between stops $\sum\limits_{i} TT_i + DT_i$. 

Dwell time is defined as the time spent within the GPS radius of a stop, including door open time. Dwell time is a combination of a part that is driven within the radius and a part that is stopped in the radius. We assume that the dwell time is mostly determined by the door open time. We formulate dwell time as the sum of door open time and add a random value drawn from the historic distribution when no passengers got into the bus at this stop. This distribution holds information about actual drive time of the GPS radius. 

This combines to Equation~\ref{eq:dwelltimefactors} 
\begin{equation}
DT_i = DOT_i + DT_i(0)
\label{eq:dwelltimefactors}
\end{equation}
where:\vspace{0.1cm}\\
$DT_i$: dwell time at stop $i$\vspace{0.1cm}\\
$DOT_i$: door open time at stop $i$\vspace{0.1cm}\\
$DT_i(0)$: dwell time when no passengers got in at stop $i$\vspace{0.1cm}\\

Door open time is again mainly determined by passengers boarding and alighting. We have determined the door open time at all intermediate stops as can be seen in Equation~\ref{eq:dooropenfactors}.

\begin{equation}
DOT_i = 6.4 + \max(2.8\cdot CI_i; 1.3\cdot CO_i)
\label{eq:dooropenfactors}
\end{equation}
where:\vspace{0.1cm}\\
$DOT_i$: door open time at stop $i$\vspace{0.1cm}\\
$CI_i$: n.o. passengers got that in at stop $i$\vspace{0.1cm}\\
$CO_i$: n.o. passengers got that off at stop $i$\vspace{0.1cm}\\

Factors are determined by ordinary least squares linear regression. For most stops, the difference in parameters of the regressors are minimal, which is why we choose to use a single linear model for all stops. The exceptions are the terminal stops at which we just sample from the historical departure time. 

\begin{figure}[h!]
	\begin{subfigure}{0.475\textwidth}
		\includegraphics[width=1\linewidth]{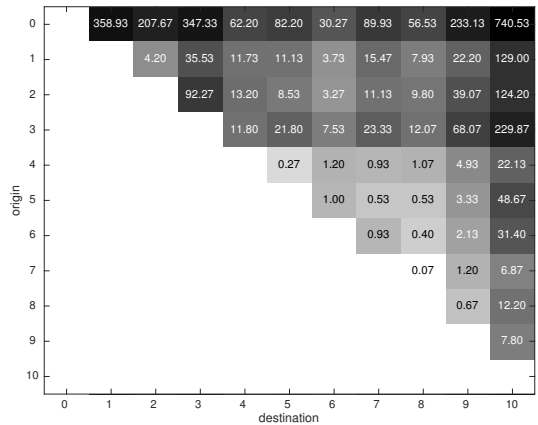}
		\caption{O-D Matrix A-direction}
	\end{subfigure}
	\begin{subfigure}{0.475\textwidth}
		\includegraphics[width=1\linewidth]{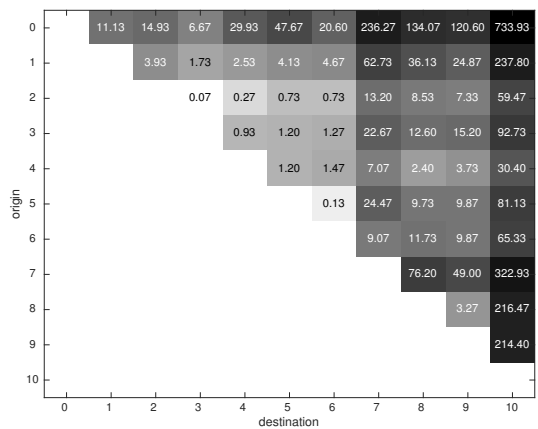}
		\caption{O-D Matrix B-direction}
	\end{subfigure}

	\caption{Average daily Origin-Destination (O-D) Matrices for both route directions. O-D Matrices give the number of passenger that travel between different stops.}
	\label{fig:OD-matrix-original}
\end{figure}

We already discussed the passenger arrival distribution at a stop for a specific trip, but not the number of passengers that actually arrive at the stop for that trip. The sum of passenger travels during a day can be specified as an origin-destination (O-D) matrix. For our specific line, the average O-D matrix for a day is shown in Figure~\ref{fig:OD-matrix-original}.

We know the check-in and check-out distribution of passengers over the stops from these matrices and we also know the number of check-ins for every combination of stops and trip id from our data. Combining this information, passengers can be accurately generated.

To show the model can be used to analyse cases with a change in the number of passengers, an Excel spreadsheet in which the increase or decrease of passengers can be specified is created. This way, different predictions on passenger growth can be easily evaluated and their effect on several KPIs will become apparent.

In this spreadsheet, a global in/decrease of passengers can be set. This can also be specified for each direction of the route. Moreover, if we really want to be specific, an increase for a scheduled trip in number of passengers can be specified. Each of these specifications can be narrowed down to specific stops, such that the change only effects these stops.

This is useful, since there might be cases where some stops may have a larger growth, e.g. due to a new company close to the stop. These passengers may also travel specifically to another stop or spread out over the subsequent stop according to the historical distribution. 

These factors alter the underlying O-D matrices on which the generation of passengers is based.

By making these options available in an easy to use manner, policy makers are able to analyse different scenarios and make solid conclusions about passenger growth.

\subsection{Visualization}
To visualize the results of our model, we have created two options. The first option is a single simulation in which each modelled agent can be seen on a geographic (GIS) map, based on OpenStreetMap\footnote{OpenStreetMap, https://www.openstreetmap.org/, Visited: Feb 22,2017}. This simulation runs a single working day and the user can specify the speed of the simulation between $\frac{1}{4}$-th of real-time and 500$\times$ real-time. This visualization allows policy makers to see in detail the effects that some of their policies have and where problems occur. A screenshot of this visualization can be seen in Appendix~\ref{subsec:appendixA}.

The second visualization option is more of a meta-visualization. This visualization is a dashboard of KPIs. The indicators are chosen in accordance to what policy makers find useful. The main indicators are: punctuality, occupancy and regularity. For each of these indicators, an overview can be seen per stop. Moreover, a detailed view for each of the stops can also be viewed to show behavioural trends during the day.
Furthermore, the average number of passengers, dwell time and travel time per stop are shown. These are used to validate our model, which we discuss in Section~\ref{sec:experiments}. 
 
\subsection{Model assumptions and limitations}
Several assumptions are made to create the simulation model. These assumptions are listed here. Details about why they are made are discussed, including their effect on accuracy of the model.

Our first assumption is that all passengers in the data represent all passengers that travel by bus. Apart from errors in the chipcard system on which the counts are based, passengers still have the possibility of buying a ticket at the bus driver. This is not registered by the system and does affect several of our models' variables. Moreover, a person buying a ticket most likely has a larger effect on dwell time than a person using the automated chipcard. 
However, from revenue statistics is shown that the number of people that buy a ticket on this specific bus line is less than 1\%. We also performed a small case study during rush hour to count how many people bought a ticket. Between 6AM and 11AM on two separate mornings, only 1 person bought a ticket on the bus. Since this is such a small effect on total statistics, we choose to disregard it.

Since we only know the number of passengers that got in at a specific trip, we had to make an assumption of passenger arrival rates. From literature, we assume that on this high-frequency line, the number of passengers that arrive is Poisson distributed. We have to note, however, that in off peak hours the schedule is changed and the longer the scheduled headway between buses is, the less likely this assumption holds. People may start to schedule their arrival more. Furthermore, this assumption does not directly use external factors, such as a train arriving with lots of transit passengers, or work schedules of commuters. 

For our passenger growth model, we allow to specify a certain growth percentage on different stops or even different trips. However, we do not allow to specify growth between different times of days, other than specifying growth for several specific trips. Some passengers may change their behaviour according to several factors. If certain bus trips approach their capacity limit, some of the passengers may change their mode of transportation or the time of day at which they travel. These effects are hard to account for, and more research is needed to see whether this may actually happen and at what scale.

Some assumptions of our travel and dwell times calculation are made. As mentioned, travel times are only dependent on the specific trip and no dependence between subsequent trip segments or trips is modelled. In reality, there is a strong association between travel times of subsequent trips and also of trip segments. If we want to model this assumption by drawing from our historical distributions we do not have enough data. To do this we would have to create a general travel time model. This is unfortunately very difficult as many external factors contribute to travel time which are not all captured within (public) data sources.
 
Another assumption that is made is that dwell time is mainly determined by the time the door opens. The deceleration and acceleration of the bus are assumed to be negligible factors, but in fact they are contributing to dwell time. Moreover, the bus may still accelerate or slow down outside the GPS radius of the stop, which may be represented within our travel times. 

Also, the door open time is linearly determined by the number of check-ins and check-outs. There is a quite high variability between passengers which we did not account for. This variability may have different causes, such as passenger behaviour, bus driver behaviour or the layout of the stop. We could also think that the time of day may affect passenger behaviour as commuters during rush hour may be more hastily than passengers off-peak. We did not take this into account. Also, we assumed that there is no difference between bus stops that contributes to dwell time. Part of the R-net concept is not only to have special buses, but also for example the design of bus stops. Bus stops are designed in a uniform matter. Not all bus stops on this line are R-net designed bus stops, but most of them are. This is why we assumed no differences between bus stops in this linear door time model.

\section{Experiments}
In this section several cases that are analysed from our model are described. First, the base model is validated using numerical analysis on dwell time, travel time, punctuality, occupancy and regularity. Thereafter, two passenger growth scenarios are simulated and their effect on different KPIs is shown.

\label{sec:experiments}
\subsection{Model validation}
First, metrics of the base model are compared to numerically calculated solutions. The most important factors are travel time and dwell time as they affect all other KPIs directly. After that, punctuality, occupancy and regularity are compared. Since there are no measurements on waiting time, a Poisson distributed arrival rate is assumed for passengers as described earlier. For waiting time therefore no comparison can be made with the numerical analysis of the data.
Comparisons made are based on means taken over 100 simulation runs versus means computed from the data.

\paragraph{Travel time}
We first compare travel time. We have computed average travel time for each trip segment in seconds for both directions and listed results by stop number in Table~\ref{tab:comptraveltime}.

\begin{table}[h!]
	\centering
	\small
	\caption{Comparison between simulation and numerical means of traveltime to each of the stops in both directions from the previous stop. Means of the model are taken over 100 simulation runs. Data means are taken over all data.}
	\label{tab:comptraveltime}
	\begin{tabularx}{0.475\textwidth}{cRRRRR}
		direction & stopnr & data (s) & model (s) & $\mid$ diff $\mid$ (s)  & $\mid$ diff $\mid$ (\%) \\\hline
		& & & & &\\
		A & 0 & NA & NA & NA & NA\\
		A & 1 & 213.83 & 213.31 & 0.52 & 0.24\\
		A & 2 & 28.09 & 28.63 & 0.55 & 1.95\\
		A & 3 & 111.69 & 112.60 & 0.91 & 0.81\\
		A & 4 & 240.01 & 239.6 & 0.39 & 0.16\\
		A & 5 & 52.72 & 52.33 & 0.39 & 0.74\\
		A & 6 & 182.91 & 185.29 & 2.38 & 1.30\\
		A & 7 & 138.38 & 139.47 & 1.09 & 0.79\\
		A & 8 & 37.57 & 37.46 & 0.11 & 0.28\\
		A & 9 & 42.45 & 42.63 & 0.18 & 0.42\\
		A & 10 & 89.05 & 89.25 & 0.20 & 0.22 \\
		& & & & &\\
		\textbf{AVG} &  &  &  & \textbf{0.67} & \textbf{0.69} \\\hline
		&  &  &  &  &\\
		B & 0 & NA & NA & NA & NA\\
		B & 1 & 100.26 & 100.75 & 0.49 & 0.49\\
		B & 2 & 52.07 & 52.66 & 0.63 & 1.21\\
		B & 3 & 29.86 & 29.92 & 0.06 & 0.20\\
		B & 4 & 143.76 & 143.18 & 0.58 & 0.40 \\
		B & 5 & 187.59 & 187.68 & 0.09 & 0.05\\
		B & 6 & 54.19 & 54.26 & 0.07 & 0.13\\
		B & 7 & 225.48 & 224.71 & 0.77 & 0.34\\
		B & 8 & 105.03 & 105.35 & 0.32 & 0.30\\
		B & 9 & 27.53 & 27.35 & 0.18 & 0.67\\
		B & 10 & 220.30 & 220.12 & 0.18 & 0.08\\
		& & & & &\\
		\textbf{AVG} &  &  &  & \textbf{0.34} & \textbf{0.39}\\\hline
	\end{tabularx}
\end{table}

Since travel time is sampled from the data, means are very close to the data. In fact, the simulation's means should eventually converge towards the numerical means. Since for 100 simulation runs the MAE on travel time is already within tenths of seconds, we conclude from these results that the amount of simulations chosen is acceptable to approach the data means. We do note that this in fact does not say that the model correctly samples the data at the level of an individual trip, but this is seen later when KPIs on a trip level are shown.

\paragraph{Dwell time}
Secondly, dwell time is inspected. Dwell time is constructed using the aforementioned linear model for door open time, based on passengers boarding and alighting and a distribution of dwell time when no passengers got in. The dwell time means per stop are shown in Table~\ref{tab:compdwelltime}. 

\begin{table}[h!]
	\centering
	\small
	\caption{Comparison between simulation and numerical means of dwell time for each of the stops. Means of the model are taken over 100 simulation runs. Data means are taken over all data.}
	\label{tab:compdwelltime}
	\begin{tabularx}{0.475\textwidth}{cRRRRR}
		direction & stopnr & data (s) & model (s) & $\mid$ diff $\mid$ (s) & $\mid$ diff $\mid$ (\%)\\\hline
		& & & & &\\
		A & 0  & NA    & NA    & NA    & NA\\
		A & 1  & 61.19 & 59.76 & 1.43  & 7.21\\
		A & 2  & 50.71 & 54.26 & 3.55  & 0.16\\
		A & 3  & 52.28 & 55.16 & 2.88  & 0.96\\
		A & 4  & 25.59 & 29.97 & 4.38  & 14.56\\
		A & 5  & 29.08 & 30.91 & 1.84  & 1.28\\
		A & 6  & 29.32 & 30.65 & 1.33  & 2.01\\
		A & 7  & 28.78 & 32.04 & 3.25  & 1.72\\
		A & 8  & 18.80 & 28.77 & 9.97  & 30.92\\
		A & 9  & 31.08 & 44.79 & 13.71 & 39.08\\
		A & 10 & NA    & NA    & NA    & NA \\
		& & & & & \\
		\textbf{AVG} &  &  &  & \textbf{4.70} & \textbf{10.88} \\\hline
		& & & & & \\
		B & 0  & NA    & NA    & NA    & NA \\
		B & 1  & 46.17 & 55.06 & 8.89  & 16.97\\
		B & 2  & 31.26 & 34.53 & 3.27  & 9.78\\
		B & 3  & 31.10 & 32.60 & 1.50  & 2.05\\
		B & 4  & 28.49 & 26.41 & 2.08  & 14.77\\
		B & 5  & 32.43 & 31.91 & 0.52  & 7.46\\
		B & 6  & 33.62 & 33.86 & 0.24  & 4.25\\
		B & 7  & 49.26 & 55.92 & 6.66  & 2.64\\
		B & 8  & 50.35 & 50.23 & 0.12  & 6.04\\
		B & 9  & 59.05 & 52.81 & 6.24  & 15.71\\
		B & 10 & NA    & NA    & NA    & NA\\
		& & & & & \\
		\textbf{AVG} &  &  &  & \textbf{3.28} &  \textbf{8.85} \\\hline
	\end{tabularx}
\end{table}

The mean absolute error (MAE) taken over stop averages, is in a different order of magnitude for dwell time than for travel time. The travel time MAE over all stops is only 0.67 and 0.34 seconds over the A and B-direction respectively, where average dwell time is 4.70 and 3.28 seconds. The same is seen from the relative error. The mean relative error is 0.69\% and 0.34\% for travel time and 10.88\% and 8.85\% for dwell time. 

Interestingly, in most cases dwell time is overestimated, but not in all cases. Since most of the door open times have a right-skewed distribution, the linear model might slightly overestimate most dwell times as it uses least squares regression on the means and not the medians. This does however not explain why the model underestimates dwell time in some cases. On the contrary, braking and accelerating are not taken into account, which could mean dwell time is underestimated. A combination of these factors might explain the difference, but also other factors might cause variations. Also, the assumption that there is no difference in dwell time between stops might not completely hold. Different traffic conditions around the stop might affect dwell time and GPS radii might not be accurately defined.

However, even though differences on dwell time are significantly higher (several seconds) than travel time we are convinced it is still accurate enough to create a truthful model, which can give actionable insight.

\paragraph{Punctuality}
Dwell time and travel time determine the performance of the KPIs. It is therefore interesting to look at the performance of these indicators. The first indicator we look at is punctuality. For this indicator, means per stop are compared. This is shown in Table~\ref{tab:comppunctuality}. Punctuality here is defined as the deviation from the scheduled departure time in seconds.

\begin{table}[h!]
	\centering
	\small
	\caption{Comparison between simulation and numerical means of punctuality for each of the stops. Means of the model are taken over 100 simulation runs. Data means are taken over all data.}
	\label{tab:comppunctuality}
	\begin{tabularx}{0.475\textwidth}{cRRRR}
		direction & stopnr & data (s) & model (s) & $\mid$ diff $\mid$ (s) \\\hline
		& & & & \\
		A & 0  & 139.17 & 149.71 & 10.54 \\
		A & 1  & 125.42 & 133.50 & 8.08  \\
		A & 2  & 144.20 & 155.74 & 11.54 \\
		A & 3  & 128.65 & 143.12 & 14.47 \\
		A & 4  & 94.64  & 113.07 & 18.43 \\
		A & 5  & 116.29 & 137.07 & 20.78 \\
		A & 6  & 88.77  & 110.99 & 22.22 \\
		A & 7  & 76.21  & 101.41 & 25.20 \\
		A & 8  & 72.73  & 107.75 & 35.03 \\
		A & 9  & 86.03  & 134.96 & 48.93 \\
		A & 10 & NA     & NA     & NA    \\
		& & & & \\
		\textbf{MAE} &  &  &  & \textbf{21.52} \\
		\hline
		& & & & \\
		B & 0  & 86.66  & 96.29  & 9.62  \\
		B & 1  & 64.82  & 82.91  & 18.09 \\
		B & 2  & 87.73  & 109.28 & 21.55 \\
		B & 3  & 88.36  & 111.63 & 23.26 \\
		B & 4  & 80.64  & 101.50 & 20.86 \\
		B & 5  & 60.75  & 81.41  & 20.66 \\
		B & 6  & 88.25  & 109.50 & 21.25 \\
		B & 7  & 74.91  & 102.29 & 27.38 \\
		B & 8  & 49.34  & 77.53  & 28.19 \\
		B & 9  & 28.07  & 50.15  & 22.07 \\
		B & 10 & NA     & NA     & NA   \\
		& & & & \\
		\textbf{MAE} &  &  &  & \textbf{21.29}\\
		\hline
	\end{tabularx}

\end{table}

Punctuality is determined by the accumulated sum of travel time and dwell time. simulations' average travel time is almost equal to the data's average travel time, we know that the differences between punctuality from the data and the model happen mainly because of the variations in dwell time. We can also see this from the data. Dwell time variation (and a small difference in travel time) is accumulated and shown in both directions.

However, another factor is at play, which is the departure punctuality at the first stop of the trip. Since the bus departs there, departure punctuality is sampled from the data just like travel time, if the bus is early. This is because the linear model does not apply directly for this first stop as many other factors, such as bus driver behaviour, influences dwell time here. Some bus drivers choose to load passengers earlier than others. There is too much variance when the bus is too early.
This method does however cause an offset in departure punctuality at the first stop in both directions.

\begin{figure}[h!]
	\centering
	\begin{subfigure}{0.475\textwidth}
		\includegraphics[width=1\linewidth,height=0.5\linewidth]{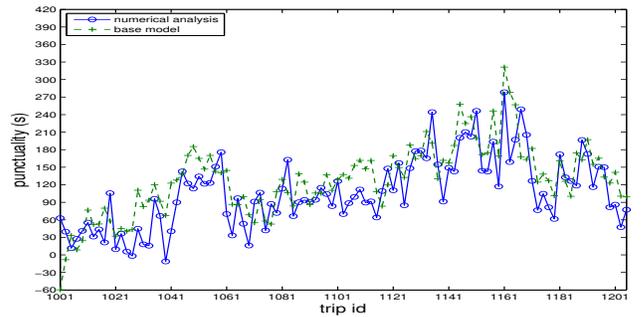}
		\caption{Punctuality A-direction}
	\end{subfigure}
	\begin{subfigure}{0.475\textwidth}
		\includegraphics[width=1\linewidth,height=0.5\linewidth]{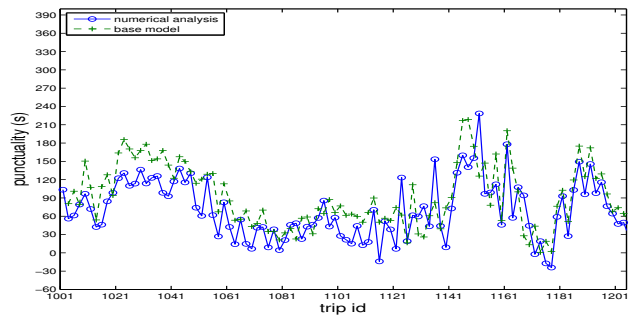}
		\caption{Punctuality B-direction}
	\end{subfigure}
	
	\caption{Average punctuality over all stops in seconds of 100 simulation runs by trip id vs numerical means of punctuality by trip id.}
	\label{fig:comppunctuality}
\end{figure}

It is useful to look at average data for each stop, but we can also aggregate data for each trip, averaging the data for all stops on that trip. By comparing trips, punctuality fluctuations over the day are shown. This is shown in Figure~\ref{fig:comppunctuality}.

What is seen from this figure is that punctuality does follow the same general trend as the data over the day, but when looking at the level of individual trips, a lot of variation between the simulation is seen. In general, punctuality is overestimated, due to the effect of dwell time overestimation. This is something we do have to take into account when interpreting the model.

What policy makers can gain from this figure is a general trend over the day. There are high peaks during rush-hour (which can be identified, since trip id's are incremental by time of day), but also there seems to be a peak in both directions somewhat later in the evening. 

\paragraph{Occupancy}
The next indicator is occupancy. Occupancy measures how many passengers (pax) there are in the bus at the point that the bus departs from a stop. It is the summed difference of all check-ins and check-outs for that trip, up till and including the stop. Passengers are generated based on the data for a scheduled trip time. If the bus is too early or too late, they might take another bus, which decreases the occupancy in one bus, but increases it in the other. Large punctuality variations between trips, which are analysed later as regularity, might directly affect occupancy in this way. 

In Table~\ref{tab:compoccupancy}, the average occupancy after each of the stops is compared to the data. 

\begin{table}[h!]
	\centering
	\small
	\caption{Comparison between simulation and numerical means of bus occupancy for each of the stops. Means of the model are taken over 100 simulation runs. Data means are taken over all data.}
	\label{tab:compoccupancy}
	\begin{tabularx}{0.475\textwidth}{cRRRR}
		direction & stopnr & data (pax) & model (pax) & $\mid$ diff $\mid$  (pax) \\\hline
		& & & & \\
		A & 0 & 21.32 & 21.36 & 0.04 \\
		A & 1 & 20.14 & 21.14 & 1.00 \\
		A & 2 & 21.00 & 22.46 & 1.46 \\
		A & 3 & 20.15 & 22.20 & 2.05 \\
		A & 4 & 19.51 & 21.51 & 2.00 \\
		A & 5 & 18.87 & 20.80 & 1.93 \\
		A & 6 & 18.74 & 20.65 & 1.91 \\
		A & 7 & 17.42 & 19.25 & 1.83 \\
		A & 8 & 16.71 & 18.47 & 1.77 \\
		A & 9 & 13.16 & 14.50 & 1.34 \\
		A & 10 & NA & NA & NA \\
		& & & & \\
		\textbf{MAE} &  &  &  & \textbf{1.53}\\
		\hline
		& & & & \\
		B & 0 & 13.11 & 13.09 & 0.02 \\
		B & 1 & 16.71 & 16.76 & 0.05 \\
		B & 2 & 17.46 & 17.58 & 0.12 \\
		B & 3 & 18.80 & 18.96 & 0.16 \\
		B & 4 & 18.93 & 19.22 & 0.29 \\
		B & 5 & 19.63 & 20.10 & 0.47 \\
		B & 6 & 20.28 & 20.85 & 0.57 \\
		B & 7 & 20.97 & 21.37 & 0.39 \\
		B & 8 & 20.30 & 20.91 & 0.61 \\
		B & 9 & 20.01 & 20.58 & 0.57 \\
		B & 10 & NA & NA & NA \\
		& & & & \\
		\textbf{MAE} &  &  &  & \textbf{0.32}\\
		\hline
	\end{tabularx}
\end{table}

\begin{figure}[h!]
	\centering
	\begin{subfigure}{0.475\textwidth}
		\includegraphics[width=1\linewidth,height=0.5\linewidth]{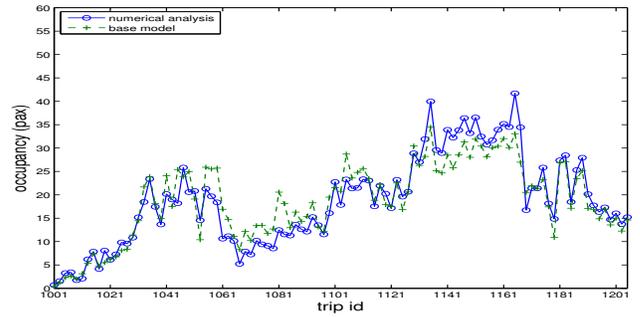}
		\caption{Occupancy A-direction}
	\end{subfigure}
	\begin{subfigure}{0.475\textwidth}
		\includegraphics[width=1\linewidth,height=0.5\linewidth]{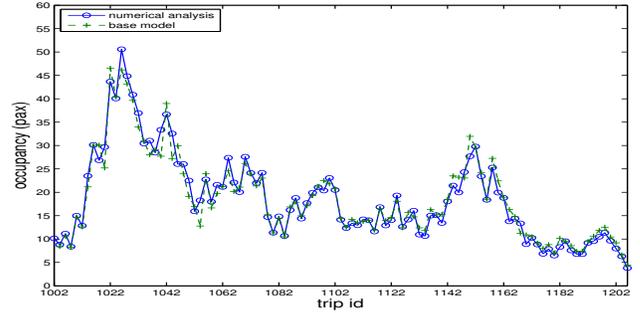}
		\caption{Occupancy B-direction}
	\end{subfigure}
	
	\caption{Average number of passengers in bus (pax) of 100 simulation runs over all stops by trip id vs numerical means.}
	\label{fig:compoccupancy}
\end{figure}

There is only a minor difference between the data and the simulation. This does say that passengers are generated correctly, but since the data is averaged, we can not directly see if passengers get on the bus that they are scheduled for. It is therefore again useful to look at the bus occupancy for each trip. Not only can trips that are close to capacity be seen, this can later on be related to variations in regularity. The occupancy over the day is shown in Figure~\ref{fig:compoccupancy}.

Again, over the day, bus occupancy in the simulation is very close to the data, even on a trip level. We can therefore conclude that, even though there are some differences between the data and simulation's punctuality, there are no large differences between occupancy of trips. 

These graphs are again also interesting for policy makers to see general trends over the day. The occupancy is obviously high during rush-hours. These buses have 44 seats, so what we see is that, even on average, the bus is already close to capacity on some trips.

\paragraph{Regularity}
Regularity says something about the difference between the scheduled headway and the actual headway. The larger this difference, the more some buses will be under or overcapacity. We see that there are some variations in punctuality with the data, but that occupancy does not seem to suffer from that. It is therefore interesting to look at regularity. We calculate regularity from the time a bus departed from a stop and the time the next bus departed from that stop. If all buses are delayed for the same amount of time, regularity will not suffer. However, if one bus is delayed more than the next, it will have an effect on regularity. Average results are shown in Table~\ref{tab:compregularity}.

\begin{table}[h!]
	\centering
	\caption{Comparison between simulation and numerical means of regularity for each of the stops. Means of the model are taken over 100 simulation runs. Data means are taken over all data.}
	\label{tab:compregularity}
	\begin{tabularx}{0.475\textwidth}{cRRRR}
		direction & stopnr & data & model & $\mid$ diff $\mid$ \\\hline
		& & & & \\
		A & 0 & 0.196 & 0.247 & 0.051 \\
		A & 1 & 0.205 & 0.267 & 0.062 \\
		A & 2 & 0.213 & 0.275 & 0.062 \\
		A & 3 & 0.213 & 0.297 & 0.084 \\
		A & 4 & 0.221 & 0.327 & 0.106 \\
		A & 5 & 0.228 & 0.336 & 0.107 \\
		A & 6 & 0.231 & 0.349 & 0.118 \\
		A & 7 & 0.230 & 0.356 & 0.117 \\
		A & 8 & 0.242 & 0.362 & 0.120 \\
		A & 9 & 0.247 & 0.369 & 0.122 \\
		A & 10 & NA & NA & NA \\
		& & & & \\
		\textbf{MAE} &  &  &  & \textbf{0.095}\\
		\hline
		& & & & \\
		B & 0 & 0.147 & 0.247 & 0.099 \\
		B & 1 & 0.164 & 0.268 & 0.104 \\
		B & 2 & 0.175 & 0.276 & 0.101 \\
		B & 3 & 0.185 & 0.298 & 0.112 \\
		B & 4 & 0.193 & 0.328 & 0.135 \\
		B & 5 & 0.203 & 0.337 & 0.134 \\
		B & 6 & 0.211 & 0.350 & 0.139 \\
		B & 7 & 0.238 & 0.359 & 0.120 \\
		B & 8 & 0.251 & 0.364 & 0.113 \\
		B & 9 & 0.258 & 0.373 & 0.114 \\
		B & 10 & NA & NA & NA \\
		& & & & \\
		\textbf{MAE} &  &  &  & \textbf{0.117}\\
		\hline
	\end{tabularx}
\end{table}

\begin{figure}[h!]
	\centering
	\begin{subfigure}{0.475\textwidth}
		\includegraphics[width=1\linewidth,height=0.5\linewidth]{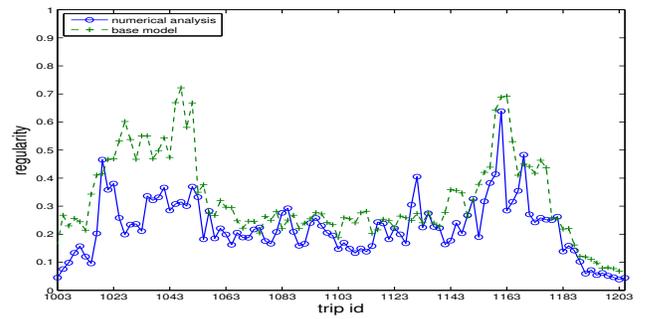}
		\caption{Regularity A-direction}
	\end{subfigure}
	\begin{subfigure}{0.475\textwidth}
		\includegraphics[width=1\linewidth,height=0.5\linewidth]{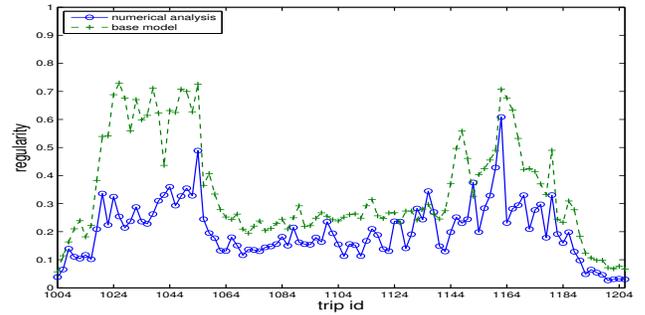}
		\caption{Regularity B-direction}
	\end{subfigure}
	
	\caption{Average regularity over all stops by trip id of 100 simulation runs vs numerical means.}
	\label{fig:compregularity}
\end{figure}

We see that there is an overestimation of regularity. This means that in the simulation, the variations in punctuality between trips are larger than those in the data. The scheduled headway therefore has a large difference with the actual headway variation. This difference seems to be especially prominent during rush-hours when scheduled headways are only 300 seconds. We can also see this effect in Figure~\ref{fig:compregularity}.

From this plot is seen that regularity seems to be worse in the simulation than in the actual data for all trips. Some effects may contribute to this result. Correlations between travel times of subsequent buses are not specified, whilst in reality there is a strong association between them. This means that one bus could be very fast, because of ``lucky sampling", whilst the next bus could have had slow travel times. In reality, these are most likely linked to some extension and variation in regularity therefore less. Furthermore, every small variation in our dwell time calculation might add up to a larger difference, depending on the number of passengers boarding or alighting the bus. Since these differences are especially large during rush-hour, as can be seen from Figure~\ref{fig:compoccupancy}, regularity is also directly influenced.

Nonetheless, in all of the KPIs is seen that the general trend of the data is followed. This helps us conclude that, under some assumptions, we can start to simulate scenarios of passenger growth. The base model has some limitations with respect to prediction of dwell time and as an effect, also overestimating punctuality and regularity, especially during rush hour. However, looking at occupancy, this does not seem to directly affect the bus that passengers get into significantly.

\subsection{Use case: passenger growth}
Now that the base model is validated, two realistic scenarios are simulated using the model. Two scenarios are simulated as an example, but using a general input format allows a policy maker to easily simulate all possible passenger growth scenarios, varying specific stops, trips or directions at which the growth takes place. Growth results are compared for each use case to the results of the base simulation and results are listed.

\paragraph{10\% passenger growth}
The first scenario simulates a 10\% passenger growth on this specific bus line in both directions. This is a realistic scenario as last year a growth of 16\% of passengers is already realised by this specific line\footnote{R-net~growth, https://www.zuid-holland.nl/kaart/nieuws/@13327/reizigersgroei-rnet/,Visited: Feb 23, 2017}.
The Excel spreadsheet is configured to set a growth percentage of 10\%, independent of direction, stop or trip and the effect on KPIs is seen in Table~\ref{tab:passresultsstop} in the 10\% columns. 

\begin{table}[h!]
	\tiny
	\caption{Results of two growth scenarios compared to the base model. A 10\% increase in passenger growth on all stops and a 25\% passenger growth scenario on stops up till stop 4 in the A-direction, where passengers check out at stop 4 and a 25\% passenger growth on stop 6 in the B-direction where passengers check-out at subsequent stops. Punctuality, occupancy and regularity differences per stop are shown.}
	\label{tab:passresultsstop}
	\begin{tabularx}{0.475\textwidth}{cR|rr|rr|rr}
		&  & \multicolumn{2}{c|}{punctuality diff} & \multicolumn{2}{c|}{occupancy diff} & \multicolumn{2}{c}{regularity diff} \\
		direction & stop & +10\% & +25\% & +10\% & +25\% & +10\% & +25\% \\\hline
		&  &  &  &  &  &  &  \\
		A & 0  & 0.86 & -0.10 & 2.17 & 5.27  & 0.000  & 0.005 \\
		A & 1  & -0.03 & 1.96  & 2.04 & 6.12  & 0.002  & 0.007 \\
		A & 2  & 1.38  & 4.42  & 2.09 & 6.94  & 0.001  & 0.006 \\
		A & 3  & 2.31  & 6.50  & 1.95 & 8.14  & 0.002  & 0.004 \\
		A & 4  & 2.99  & 25.53 & 1.90 & 0.19  & -0.001 & 0.007 \\
		A & 5  & 4.03  & 25.42 & 1.79 & 0.12  & -0.002 & 0.006 \\
		A & 6  & 4.27  & 24.63 & 1.80 & 0.09  & -0.002 & 0.005 \\
		A & 7  & 4.99  & 24.40 & 1.64 & 0.15  & 0.000  & 0.005 \\
		A & 8  & 5.84  & 23.86 & 1.57 & -0.96 & 0.000  & 0.005 \\
		A & 9  & 6.36  & 23.56 & 1.23 & -0.76 & 0.002  & 0.006 \\
		A & 10 & NA    & NA    & NA   & NA    & NA     & NA    \\
		& & & & & & & \\
		\multicolumn{2}{c|}{\textbf{AVG:}}  & \textbf{3.30} & \textbf{16.02} & \textbf{1.82} & \textbf{2.53} & \textbf{0.000} & \textbf{0.006} \\\hline
		&  &  &  &  &  &  &  \\
		B & 0  & 2.03  & 9.38  & 1.21 & -0.01 & 0.003  & 0.020 \\
		B & 1  & 4.24  & 9.89  & 1.60 & 0.06  & 0.001  & 0.018 \\
		B & 2  & 4.97  & 10.03 & 1.64 & 0.02  & 0.001  & 0.019 \\
		B & 3  & 5.21  & 10.14 & 1.75 & 0.01  & -0.001 & 0.019 \\
		B & 4  & 6.12  & 9.97  & 1.77 & -0.01 & 0.003  & 0.020 \\
		B & 5  & 6.35  & 9.52  & 1.83 & -0.02 & 0.003  & 0.018 \\
		B & 6  & 7.15  & 10.92 & 1.89 & 0.21  & 0.002  & 0.017 \\
		B & 7  & 9.06  & 11.37 & 1.95 & 0.15  & 0.002  & 0.016 \\
		B & 8  & 10.61 & 11.29 & 1.86 & 0.13  & 0.002  & 0.017 \\
		B & 9  & 11.14 & 10.58 & 1.84 & 0.12  & 0.002  & 0.017 \\
		B & 10 & NA    & NA    & NA   & NA    & NA     & NA \\
		& & & & & & & \\
		\multicolumn{2}{c|}{\textbf{AVG:}} & \textbf{6.69} & \textbf{10.31} & \textbf{1.74} & \textbf{0.07} & \textbf{0.002} & \textbf{0.018} \\\hline
	\end{tabularx}
\end{table}

The effect on the KPIs is immediately apparent. Punctuality increases at almost every stop in both directions. Dwell time increases linearly with the passengers due to passengers boarding and alighting. As every stop has a 10\% increase in passengers, punctuality suffers more and more. On some stops this has an even larger effect due to a larger absolute increase of passengers, which are derived from the O-D matrices seen in Figure~\ref{fig:OD-matrix-original}.
Delays propagate throughout the rest of the trip. This is of course under the assumption that the bus driver can not drive any faster to make up time. 

\begin{figure}[h!]
	\centering
	\begin{subfigure}{0.475\textwidth}
		\includegraphics[width=1\linewidth,height=0.5\linewidth]{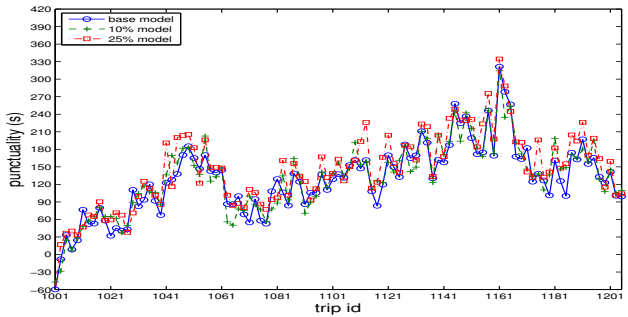}
		\caption{Punctuality A-direction}
	\end{subfigure}
	\begin{subfigure}{0.475\textwidth}
		\includegraphics[width=1\linewidth,height=0.5\linewidth]{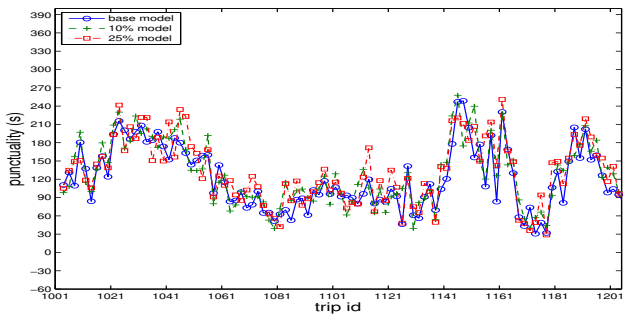}
		\caption{Punctuality B-direction}
	\end{subfigure}
	
	\caption{Average punctuality over all stops by trip id of the base model and two passenger growth scenarios. Results are averaged over 100 simulation runs.}
	\label{fig:compmodelpunctuality}
\end{figure}

This effect can also be seen when looking at punctuality by trip id as seen in Figure~\ref{fig:compmodelpunctuality}.

\begin{figure}[h!]
	\centering
	\begin{subfigure}{0.475\textwidth}
		\includegraphics[width=1\linewidth,height=0.5\linewidth]{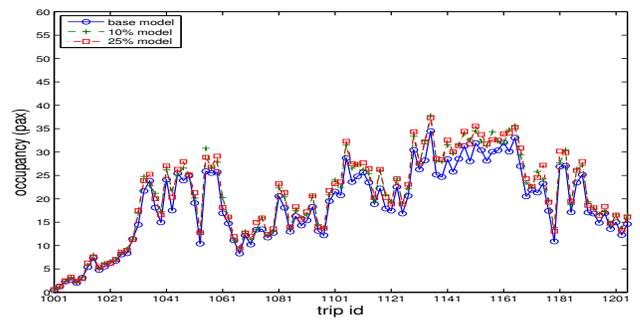}
		\caption{Occupancy A-direction}
	\end{subfigure}
	\begin{subfigure}{0.475\textwidth}
		\includegraphics[width=1\linewidth,height=0.5\linewidth]{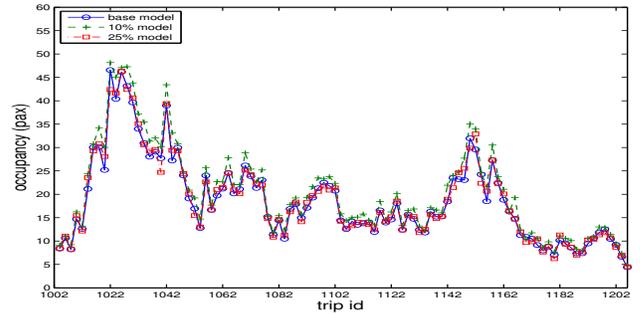}
		\caption{Occupancy B-direction}
	\end{subfigure}
	
	\caption{Average occupancy over all stops by trip id of the base model and two passenger growth scenarios. Results are averaged over 100 simulation runs.}
	\label{fig:compmodeloccupancy}
\end{figure}

For occupancy, a very clear increase over the stops can be observed. Indeed, after every stop, there are on average 10\% more passengers in the bus. In Figure~\ref{fig:compmodeloccupancy} is also seen that for almost every trip there are more passengers, on average, in the bus.

\begin{figure}[h!]
	\centering
	\begin{subfigure}{0.475\textwidth}
		\includegraphics[width=1\linewidth,height=0.5\linewidth]{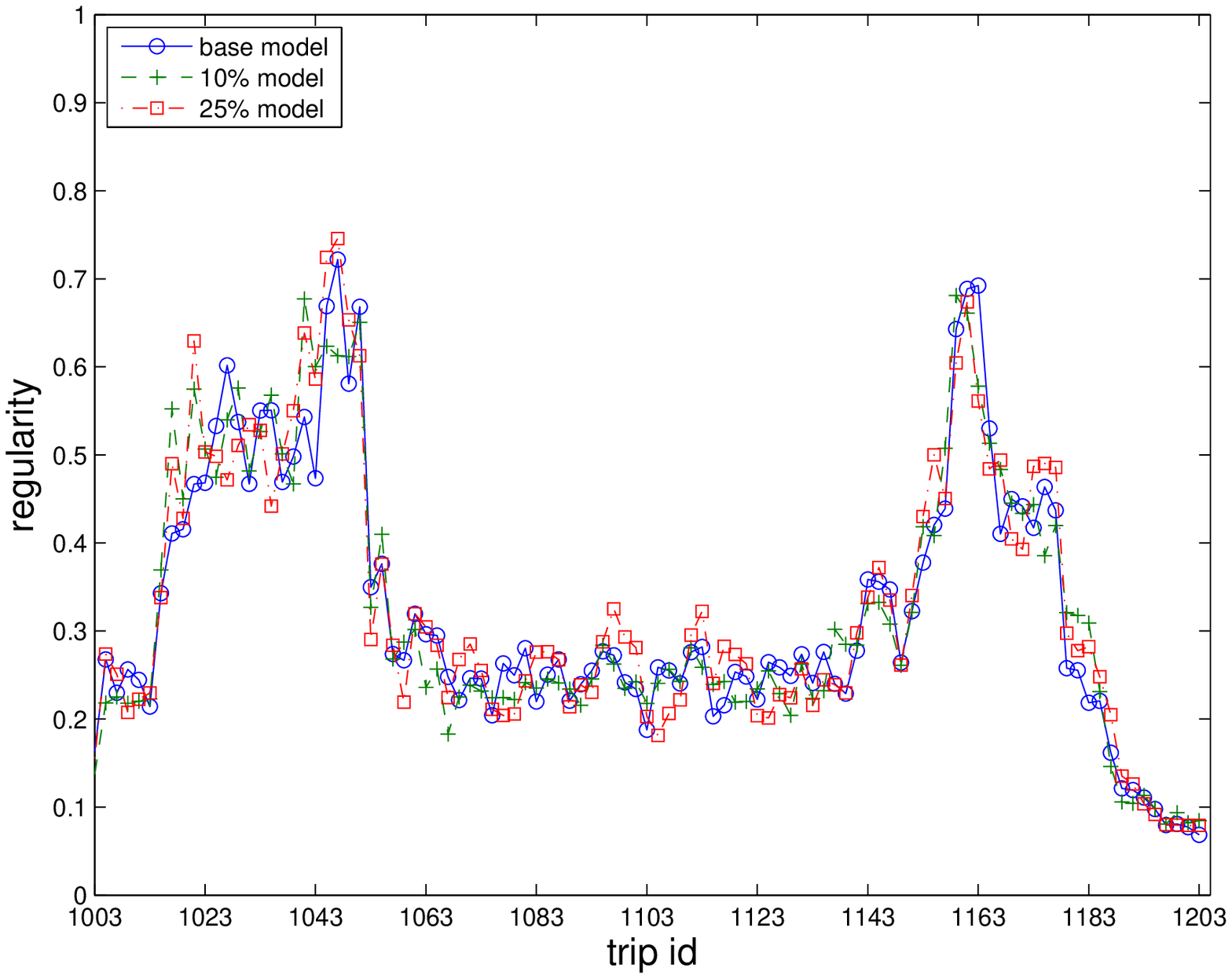}
		\caption{Regularity A-direction}
	\end{subfigure}
	\begin{subfigure}{0.475\textwidth}
		\includegraphics[width=1\linewidth,height=0.5\linewidth]{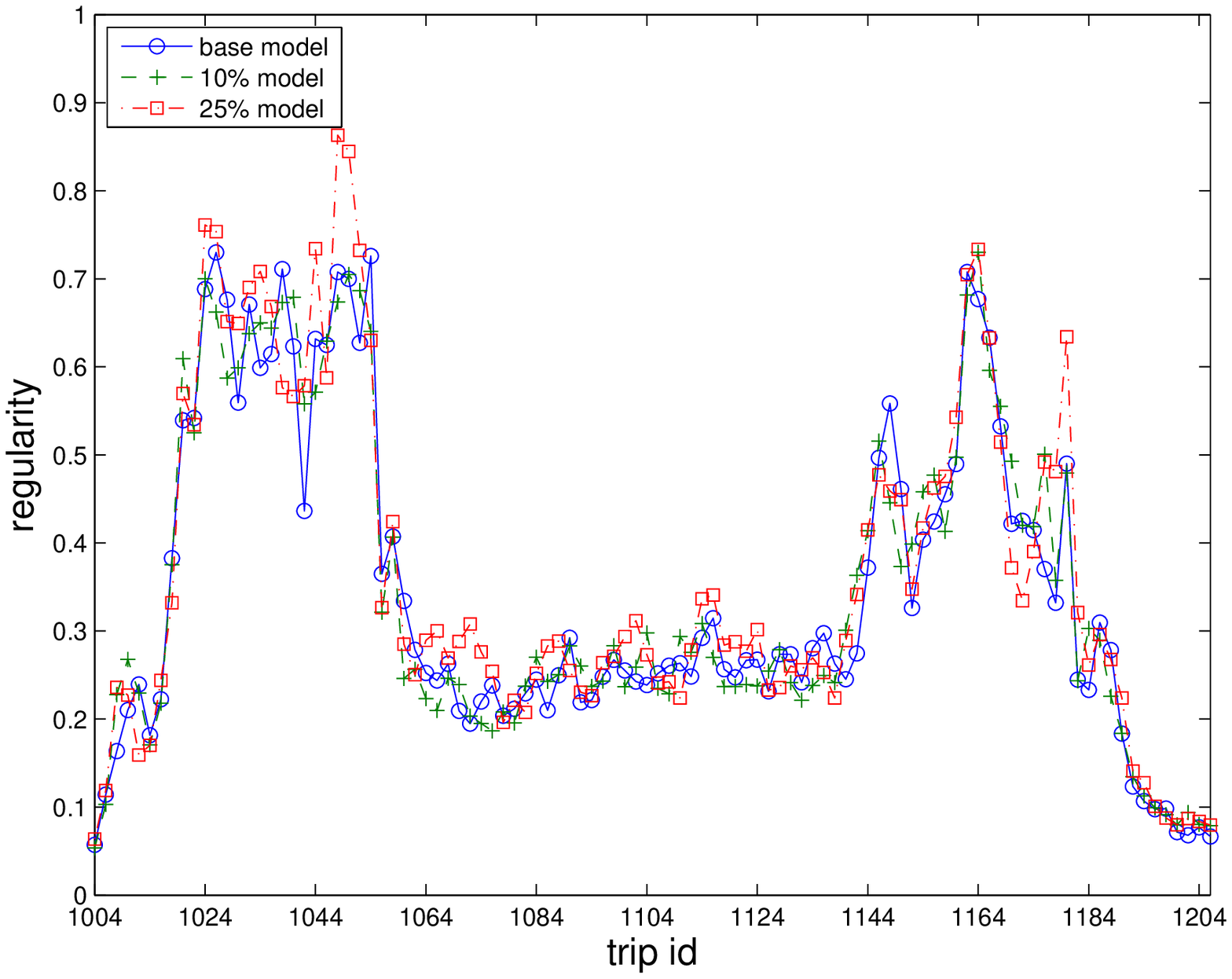}
		\caption{Regularity B-direction}
	\end{subfigure}
	
	\caption{Average regularity over all stops by trip id of the base model and two passenger growth scenarios. Results are averaged over 100 simulation runs.}
	\label{fig:compmodelregularity}
\end{figure}

The effect on regularity is less direct. We might expect that busy trips will get busier and therefore also have a larger effect on dwell time than less busier subsequent or antecedent trips. This would create a larger headway between them, which will cause regularity to suffer.
There are however other effects that take into place. The number of seats in our buses is 44, which is already exceeded for some trips without passenger growth. The model has a maximum of 60 passengers per bus. If passengers can not get into the bus, because it already is full, they will have to wait for the next bus. This might affect regularity as passengers spill over to the next bus. This means that the next bus could also be delayed more, which may have a positive effect on regularity.
Furthermore, if a bus is delayed more than its scheduled headway with the next bus, it will pick up passengers for that bus as well, which will delay it further. The effect of all these factors influence regularity. The net effect, as is seen from the average regularity, seems to be rather neutral. There are some stops that have improved regularity, but there are also some stops that have a slightly worse regularity and stops that are equal. This effect can also be seen from regularity by trip id in Figure~\ref{fig:compmodelregularity}. Some trips have a better regularity, whilst other trips have a worse regularity. In general, regularity follows the same trend as the base model and differences are nihil. 

\paragraph{25\% passenger growth on a specific stop}
The next scenario simulates a 25\% passenger growth scenario on specific stops in both directions. This again is not an unrealistically high growth  scenario as one of the stops already encountered a 51\% passenger growth in a single year. Growth percentage are set to 25\% at stops 0 to 3 in in the A-direction. All these passengers check-out at stop 4. We also increase the number of check-ins at stop 6 in the B-direction. These are opposing stops.

The city of Leiden is currently working on a parking plan, causing a radical reduction of car parking spots in the inner city. These stops are just outside the city and have a park-and-ride opportunity. People from all inner city stops may travel towards this park-and-ride and travel from it to the inner city.

The results on the KPIs are again shown in Table~\ref{tab:passresultsstop}. From this table can be seen that punctuality increases on the stops up till stop 4 in the A-direction due to people getting in the bus. Since all these people alight at stop 4, this is where punctuality mainly increases. This affects the rest of the trip as well, since the bus is already late. 

In the B-direction a general increase in punctuality is present over all stops. This may seem strange at first, since only the number of check-ins at stop 6 is increased, but since trips in the A-direction are delayed in such a way, the buses cannot depart at stop 0 in the B-direction on time. This causes an offset in punctuality. This punctuality slightly increases from stop 6 on as people alight from the bus. Since the increase in passengers is not specifically for individual trips, we see a general increase in punctuality over all trips from Figure~\ref{fig:compmodelpunctuality}.

For occupancy, the increase in the A-direction on the stops before stop 4 is seen. For the B-direction there is a small increase from stop 6 in occupancy. This seems a lot less than in the A-direction, but since a relative increase in passengers is specified and stop 6 is not a very busy stop, the average absolute increase is small. Again, the increase is for all trips as can be seen in Figure~\ref{fig:compmodeloccupancy} for individual trips. It can also be seen that busier trips seem to get even busier (the rich get richer effect), which may cause problems with regularity.
 
Regularity now seems to suffer as the procession of buses is disrupted by the increase of dwell time. Even though we only have an increase in dwell time, an average increase on all stops can be seen as this effect propagates through the system. 
Since a delay on some of the stops, due to passengers getting in, is created, the difference in headway between buses increases. The effect is however larger for already busy buses. This is because there is a difference between trip occupancy. The effect on regularity can be seen in Figure~\ref{fig:compmodelregularity}. The effect even continues in the B-direction. As we are too late in some cases for our trip in the B-direction, the effects continues in that direction as well. As the last stop in the A-direction has most alighting passengers, we start in the B-direction with an even worse regularity.

\section{Conclusions and outlook}
\label{sec:conclusions}
In this work, a hybrid agent-based and discrete event simulation model for a high-frequency bus line for the Arriva bus company is developed. 
This hybrid approach can simulate an interacting environment of buses and passengers, based on historical data. Insight is obtained in the effect on bus punctuality, occupancy and regularity. 

Different models of passenger growth can easily be analysed using the simulation and the effect on the aforementioned performance indicators are shown in a visually easily understandable manner. Two of these passenger growth scenarios are shown and their effect on the KPIs become apparent. With only a 10\% increase of passengers on all stops, on average, buses become close to capacity, especially during rush hour. Furthermore, average punctuality suffers by several seconds. With a 25\% increase at a specific stop, an even larger occupancy problem is seen on the specific route segments, especially in the A-direction towards the stop. Moreover, this causes the bus to be less punctual throughout the rest of the trip.
Being able to easily configure such scenarios helps policy makers to predict future possible bottlenecks.

The main limitation of the model is the prediction of dwell times using the current data available. The number of passengers is a limited predictor of dwell time as other factors such as acceleration, deceleration of the bus, bus stop layout and bus driver behaviour all affect dwell time. Furthermore, travel time is a composite statistic based on many factors, such as traffic lights and traffic situations. To actually predict these factors instead of drawing travel time from historical data would improve the model's predictive power.

Several assumptions on passengers are also made. This includes the  assumption that passengers in our data represent all passengers that travel by bus. Based on revenue statistics we can say that this assumption is fair.  However, the distribution of arrival rates of passengers is assumed to be Poisson distributed. This distribution may not hold in off-peak hours due to a planned arrival of passengers.

Foreseeable extensions to the model are the addition of multiple bus lines and the integration of more data. Passengers changing bus lines would be an important factor in that extension. The best allocation of buses over these bus lines, based on passenger demand can be found through that extension.
To optimize regularity, buses need to be evenly spread out over the trip. Measures to prevent bus bunching can be implemented as an extension to the model to optimize bus occupancy. Further research into the integration of measures to prevent bus bunching may help to prevent regularity.
\clearpage
%
\bibliographystyle{abbrv}
\bibliography{refs}  

\begin{thebibliography}{10}

\bibitem{bonabeau2002agent}
E.~Bonabeau.
\newblock Agent-based modeling: Methods and techniques for simulating human
  systems.
\newblock {\em Proceedings of the National Academy of Sciences}, 99(suppl
  3):7280--7287, 2002.

\bibitem{borshchev2004system}
A.~Borshchev and A.~Filippov.
\newblock From system dynamics and discrete event to practical agent based
  modeling: reasons, techniques, tools.
\newblock In {\em Proceedings of the 22nd international conference of the
  system dynamics society}, volume~22. Citeseer, 2004.

\bibitem{dessouky2003real}
M.~Dessouky, R.~Hall, L.~Zhang, and A.~Singh.
\newblock Real-time control of buses for schedule coordination at a terminal.
\newblock {\em Transportation Research Part A: Policy and Practice},
  37(2):145--164, 2003.

\bibitem{doroshenko2016evaluation}
A.~Doroshenko, W.~Qian, and N.~D. Osgood.
\newblock Evaluation of outbreak response immunization in the control of
  pertussis using agent-based modeling.
\newblock {\em PeerJ}, 4:e2337, 2016.

\bibitem{fu2002design}
L.~Fu and X.~Yang.
\newblock Design and implementation of bus-holding control strategies with
  real-time information.
\newblock {\em Transportation Research Record: Journal of the Transportation
  Research Board}, (1791):6--12, 2002.

\bibitem{hajinasab2016agenttransport}
B.~Hajinasab, P.~Davidsson, J.~A. Persson, and J.~Holmgren.
\newblock {\em Towards an Agent-Based Model of Passenger Transportation}, pages
  132--145.
\newblock Springer International Publishing, Cham, 2016.

\bibitem{jennings1999agent}
N.~R. Jennings.
\newblock Agent-based computing: Promise and perils.
\newblock In {\em Proceedings of the Sixteenth International Joint Conference
  on Artificial Intelligence}, IJCAI '99, pages 1429--1436, San Francisco, CA,
  USA, 1999. Morgan Kaufmann Publishers Inc.

\bibitem{jung2016creating}
M.~Jung, A.~B. Cla{\ss}en, and F.~Rudolph.
\newblock Creating and validating a microscopic pedestrian simulation to
  analyze an airport security checkpoint.
\newblock In {\em 2015 WINTER SIMULATION CONFERENCE PROCEEDINGS}, pages
  904--905. Omnipress, 2016.

\bibitem{mcdonnell2011exploring}
S.~McDonnell and M.~Zellner.
\newblock Exploring the effectiveness of bus rapid transit a prototype
  agent-based model of commuting behavior.
\newblock {\em Transport Policy}, 18(6):825--835, 2011.

\bibitem{ostle1975statistics}
B.~Ostle and L.~C. Malone.
\newblock {\em Statistics in research: Basic concepts and techniques for
  research workers}.
\newblock Iowa State press, 1988.

\bibitem{postema2015anylogic}
B.~F. Postema and B.~R. Haverkort.
\newblock An anylogic simulation model for power and performance analysis of
  data centres.
\newblock In {\em European Workshop on Performance Engineering}, pages
  258--272. Springer, 2015.

\bibitem{robinson2014simulation}
S.~Robinson.
\newblock {\em Simulation: the practice of model development and use}.
\newblock Palgrave Macmillan, 2014.

\bibitem{toledo2010meso}
T.~Toledo, O.~Cats, W.~Burghout, and H.~N. Koutsopoulos.
\newblock Mesoscopic simulation for transit operations.
\newblock {\em Transportation Research Part C: Emerging Technologies},
  18(6):896 -- 908, 2010.
\newblock Special issue on Transportation SimulationAdvances in Air
  Transportation Research.

\bibitem{vanoort2011service}
N.~van Oort.
\newblock {\em Service reliability and urban public transport design}.
\newblock PhD thesis, Delft University of Technology, 2011.

\end{thebibliography}
%
%
\balancecolumns
\onecolumn
\clearpage
\appendix

\section{Visualization of a single day}
\label{subsec:appendixA}
\begin{figure}[h!]
	\centering
	\includegraphics[width=0.9\linewidth,height=1\linewidth]{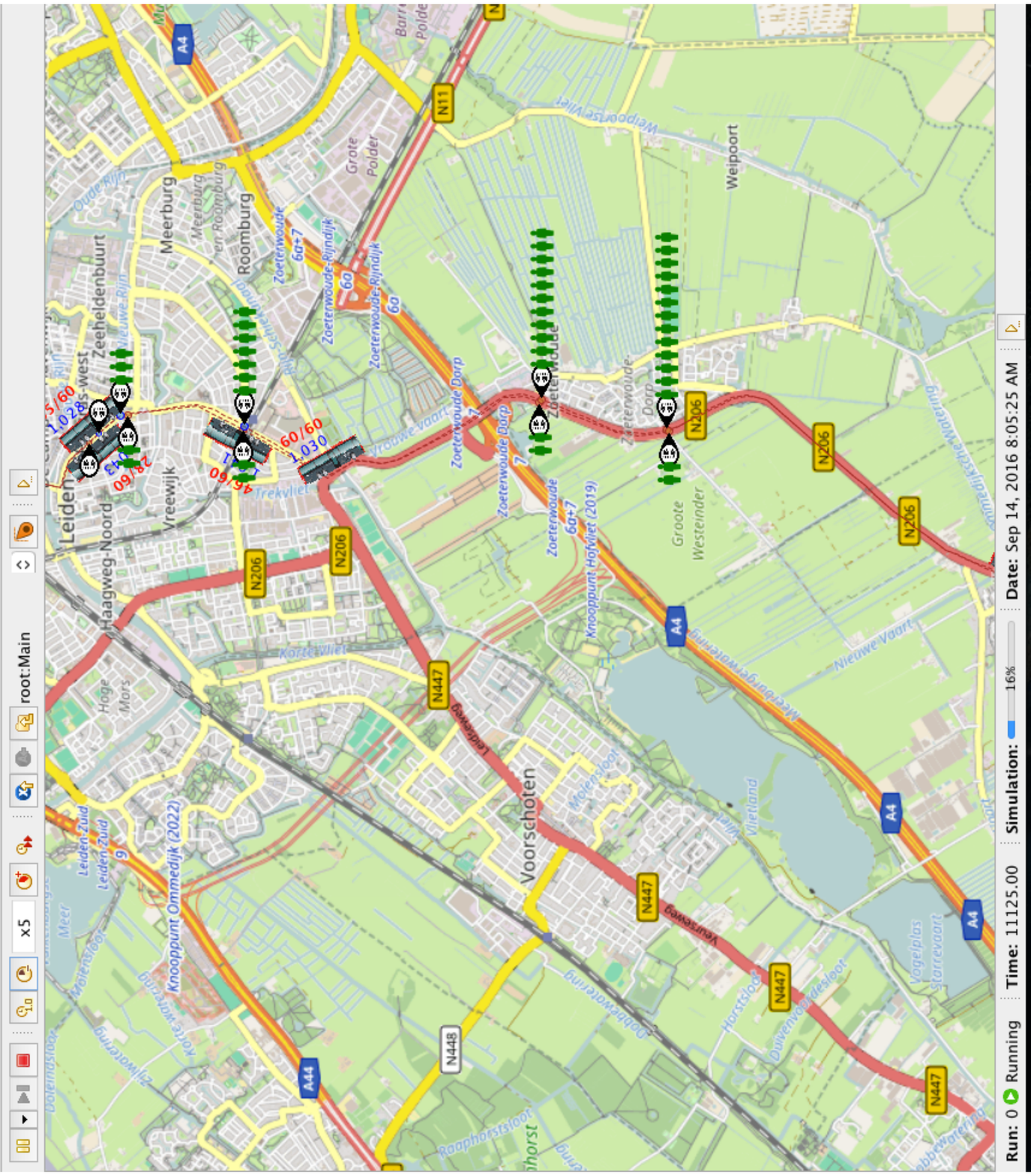}
	\caption{Visualization of a single day. Passengers can be seen next to the stops in green. Buses drive along the route. Trip id's and occupancy are shown above the buses.}
	\label{fig:singledayview}
\end{figure}
\clearpage
\section{Dashboard of multiple simulations}
\label{subsec:appendixB}
\subsection{Global overview of KPIs}
\begin{figure}[h!]
	\centering
	\includegraphics[angle=-90,width=0.7\linewidth]{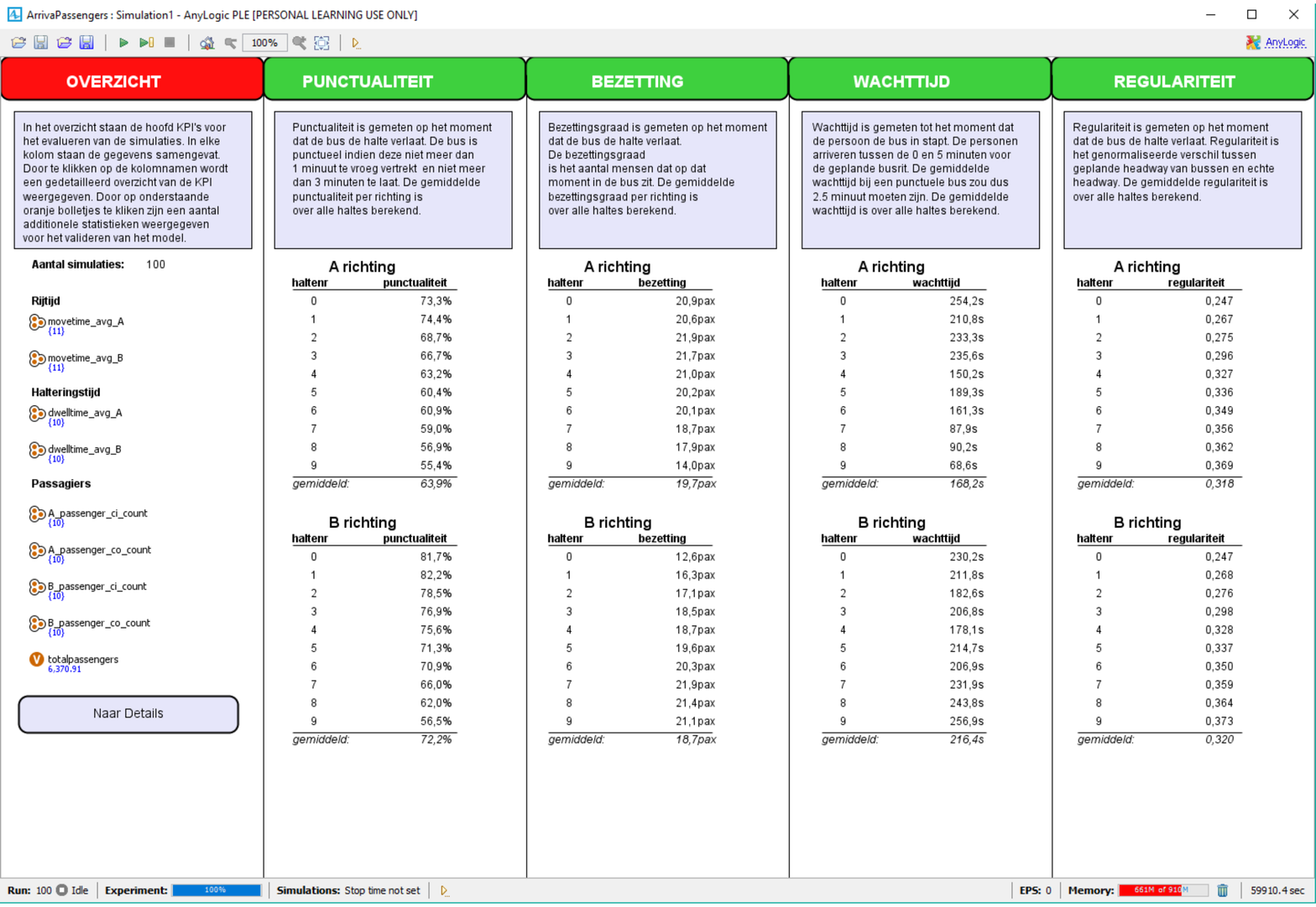}
	\caption{Dashboard overview. For multiple simulations, statistics are gathered on several KPIs and averages are shown by stop.}
	\label{fig:overview}
\end{figure}

\subsection{Detailed view of a single KPI}
\begin{figure}[h!]
	\centering
	\includegraphics[angle=-90,width=0.7\linewidth]{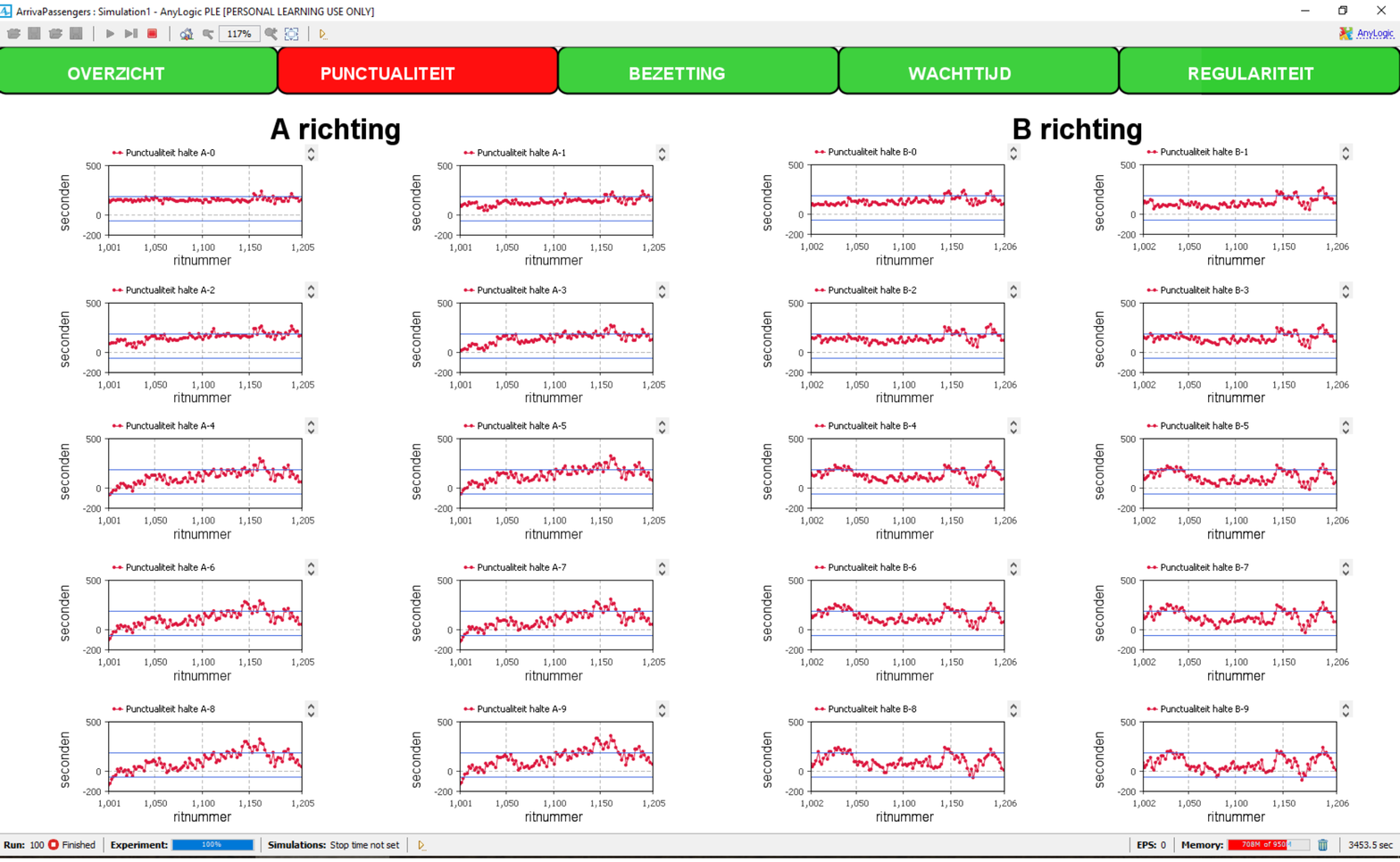}
	\caption{Dashboard of a single KPI, namely punctuality. For each stop, punctuality is shown for every trip of the day in both directions.}
	\label{fig:detailedKPI}
\end{figure}

\end{document}